\documentclass[10pt,twocolumn,letterpaper]{article}
\usepackage{stfloats}
\usepackage[pagenumbers]{cvpr} 
\usepackage{multirow}

\usepackage{colortbl}
\usepackage{float}

\usepackage{xspace}
\newcommand{\MethodName}{AnyScene\xspace}

\newcommand{\myparagraph}[1]{\vspace{0.1cm}\noindent{\bf #1}}

\definecolor{cvprblue}{rgb}{0.21,0.49,0.74}
\usepackage[pagebackref,breaklinks,colorlinks,allcolors=cvprblue]{hyperref}

\title{AnyScene: Towards Highly Controllable Driving Scene Generation \\ at Anywhere and Beyond}

\author{
Haiming Zhang$^{1*}$ \quad
Junfei Zhou$^{1,2*}$ \quad
Feng Jiang$^{1*}$ \quad
Jingzhong Li$^{1*}$ \\
Zhenglong Guo$^{1}$ \quad
Penglin Dai$^{2}$ \quad
Jifeng Dai$^{3}$ \quad
Yan Xie$^{1}$ \quad
Benjin Zhu$^{1,3\dagger}$ \\
$^{1}$Li Auto \quad
$^{2}$Southwest Jiaotong University \quad
$^{3}$Tsinghua University \\
{\tt\small
\{zhanghaiming3, zhubenjin\}@lixiang.com, jeffreychou@my.swjtu.edu.cn
} \\
{\small $^{*}$Equal contribution. \quad $^{\dagger}$Corresponding author.}\\
{\normalfont\small
    Project Page:\, \url{https://mind-omni.github.io/}}
}

\makeatletter
\g@addto@macro\@maketitle{
    \vspace{-0.5cm}
    \vspace{-10pt}
    \begin{figure}[H]
        \setlength{\linewidth}{\textwidth}
        \setlength{\hsize}{\textwidth}
        \centering
        \includegraphics[width=0.99\linewidth]{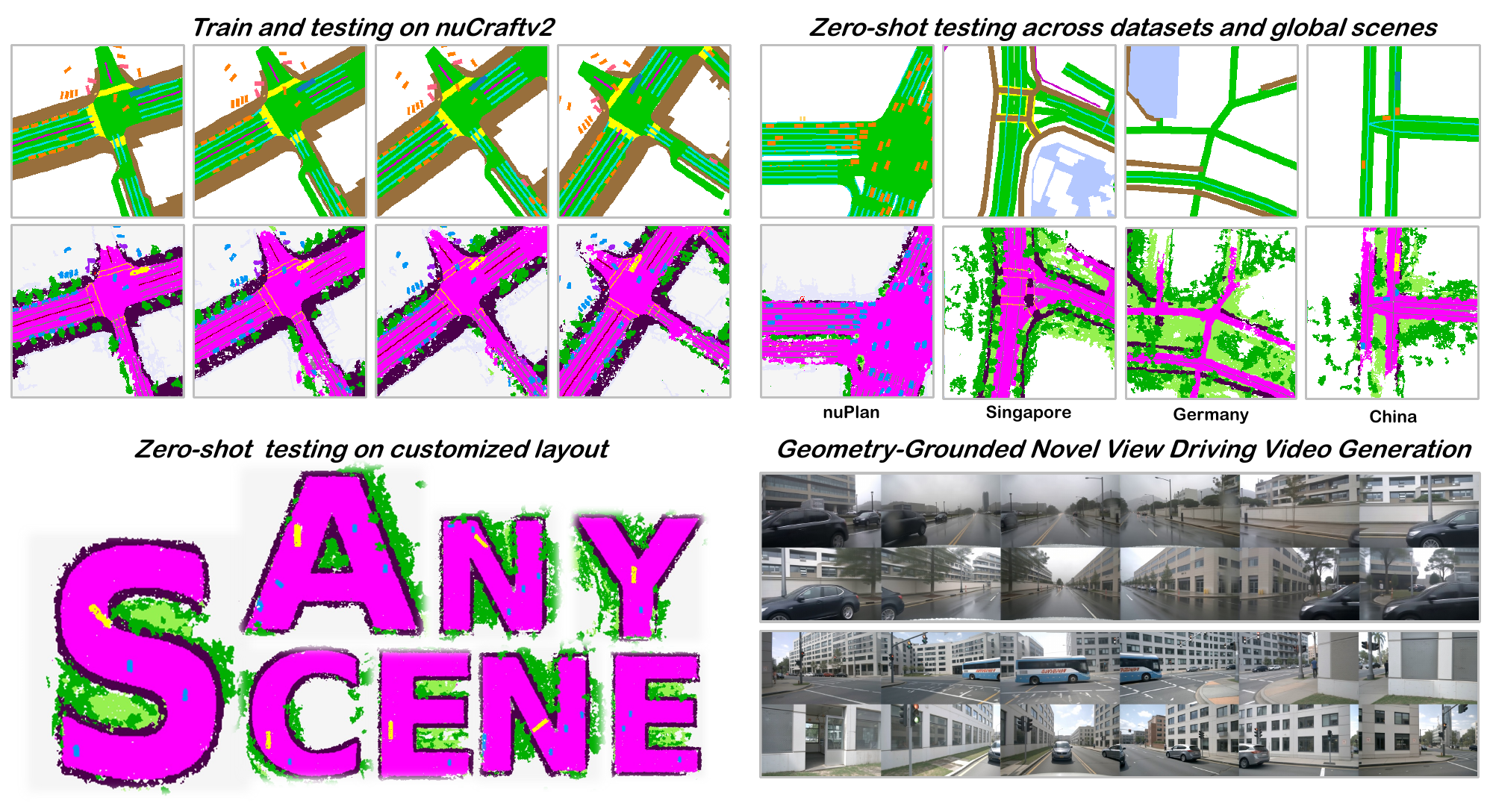}
        \vspace{-4pt}
        \caption{\textbf{An overview of the proposed \MethodName framework for driving scene generation.}
        \MethodName is trained exclusively on our created nuCraftv2 dataset and demonstrates strong zero-shot generalization to unseen datasets as well as to BEV layouts derived from OpenStreetMap (OSM)~\cite{OpenStreetMap2025} across global scenes. These OSM-derived layouts can be further augmented by manually placing agents, as shown in the China example.
        The framework also supports zero-shot occupancy generation from fully customized BEV layouts, further illustrating its high controllability.
        Additionally, the geometry-grounded view expansion module enables the synthesis of temporally consistent multi-view driving videos with an arbitrary number of camera views (\textbf{12 views} are shown here).
        }
        \vspace{-1pt}
        \label{fig:teaser}
    \end{figure}
}
\makeatother

\begin{document}
\maketitle
\begin{abstract}
Generating high-fidelity and controllable synthetic data is critical for advancing end-to-end autonomous driving, particularly for addressing the long tail of rare safety-critical scenarios.
Existing occupancy-guided methods typically rely on shallow conditioning mechanisms and reference-frame-dependent video synthesis, which limits fine-grained controllability from arbitrary BEV layouts and restricts their applicability for scalable simulation.
In this paper, we propose \textbf{\MethodName}, a unified occupancy-centric framework for driving scene generation.
\MethodName generates semantic occupancy sequences from BEV layouts through a Spatial-Temporal Occupancy Diffusion Transformer that jointly tokenizes BEV and occupancy features in an autoregressive manner.
This design enables precise controllability from cross-dataset and user-defined BEV inputs while naturally supporting long-horizon generation.
Building upon the generated occupancy, a Geometry-Grounded View Expansion module treats occupancy as the canonical spatial representation and synthesizes temporally consistent multi-view driving videos in a reference-free and autoregressive fashion, supporting flexible camera configurations at inference time.
Extensive experiments demonstrate that \MethodName achieves state-of-the-art performance in both occupancy and video generation.
It exhibits strong generalization to unseen and customized layouts, and provides measurable benefits for downstream tasks such as sparse-view 3D reconstruction.
\end{abstract}

\section{Introduction}
\label{sec:intro}

Despite substantial progress in L2+ assisted driving~\cite{hu2023planning,zhou2025autovla,li2025recogdrive,luo2026lastvla}, achieving reliable Level-4 autonomy remains hindered by the long-tail distribution of rare yet safety-critical scenarios. These scenarios occur infrequently in real-world operations and are severely underrepresented in existing public datasets~\cite{caesar2020nuscenes,sun2020scalability,argoverse2}. 
As a result, scalable synthetic data generation has become essential for systematically addressing this long-tail challenge.

To support effective closed-loop training and evaluation of end-to-end autonomous driving systems, a next-generation driving simulator must simultaneously satisfy two fundamental requirements: \emph{precise controllability}, which enables faithful instantiation of user-specified scene configurations, and \emph{anywhere-deployability}, which requires robust generalization to unseen geographies, traffic patterns, and cross-dataset or user-defined layouts. Satisfying both requirements concurrently remains a central and unresolved challenge.

Existing approaches to driving scene synthesis each exhibit structural limitations that prevent them from meeting these dual requirements. 
Direct end-to-end video generation models~\cite{hu2023gaia,gaia-2,gao2024vista,gao2025magicdrivev2,drivedreamer-2,wen2024panacea} can produce visually compelling sequences from coarse prompts, yet their implicit mappings often result in inconsistent geometry across views and time, making fine-grained control difficult. 
Reconstruction-based simulators~\cite{wu2023mars,yang2023emernerf,tonderski2024neurad,yan2024street,zhou2024drivinggaussian,chen2024omnire} achieve high visual fidelity but are inherently limited to replaying logged trajectories due to per-scene optimization. 
These shortcomings motivate the use of intermediate representations that decouple geometry from appearance.

Among such representations, 3D semantic occupancy has emerged as a promising choice because it encodes scene geometry and semantics in a structured, view-independent manner, allowing direct and editable control over scene elements~\cite{cao2022monoscene,huang2023tri,zheng2023occworld,wang2024occsora}. 
In autonomous driving, BEV layouts provide a particularly suitable conditioning signal, as they naturally align with perception and planning outputs~\cite{hu2023planning,gao2024magicdrive,uniscene}. 
Building on these advantages, recent occupancy-guided methods such as UniScene~\cite{uniscene} and InfiniCube~\cite{infinicube} have explored this direction. However, their practical capabilities remain limited in three key aspects.
First, the mapping from BEV layouts to occupancy is fragile. Existing pipelines typically rely on shallow cross-attention adapters and over-engineered triplane factorizations~\cite{geniedrive,SemCity}, which weaken fine-grained control and often lead to missing agents, blurred layout edits, or mode-collapsed attributes. 
Second, conditioning signals are typically provided at low frequency ($2$\,Hz)~\cite{caesar2020nuscenes, geniedrive, tian2023occ3d} or at coarse quality~\cite{uniscene}, both of which are incompatible with the higher frame rates required by practical video generation pipelines.
Third, multi-view video synthesis in current methods~\cite{uniscene,infinicube,geniedrive} structurally depends on reference frames and a fixed camera rig, preventing the generation of entirely new scenes from arbitrary BEV inputs and limiting flexibility in camera configuration.

To address these limitations, we propose \textbf{\MethodName}, a unified occupancy-centric framework for driving scene generation. 
Given an arbitrary BEV layout, \MethodName first produces a high-fidelity semantic occupancy sequence through a BEV-conditioned generative model. 
This model employs a compact 2D BEV-based Occupancy VAE to encode occupancy into latent representations, and a Spatio-Temporal Occupancy Diffusion Transformer (STOccDiT) that jointly tokenizes BEV layout features and occupancy latents. 
By adopting an autoregressive generation process with a causal temporal attention mask, STOccDiT enables precise and high-frequency control from arbitrary, cross-dataset, and user-defined BEV inputs while naturally supporting long-horizon occupancy sequences. 
Building upon the generated occupancy, \MethodName further synthesizes temporally and cross-view consistent multi-view driving videos through Geometry-Grounded View Expansion (GGVE). 
This module treats the occupancy sequence as the canonical spatial representation and renders explicit geometric conditioning signals to guide video generation in a reference-free and autoregressive manner, allowing flexible control over camera counts and poses at inference time. 
To support high-frequency evaluation aligned with practical video pipelines, we additionally introduce \emph{nuCraftv2}, a $12$\,Hz resampling of nuScenes with synchronized BEV layouts and high quality dense semantic occupancy.

We validate \MethodName through comprehensive experiments, demonstrating state-of-the-art performance across occupancy generation quality, multi-view video fidelity, and geometry-grounded controllability. Additional qualitative results further showcase the flexibility and versatility of \MethodName in video synthesis across arbitrary camera rigs, novel trajectories, scene editing, and scene reconstruction.

In summary, the main contributions of this work are as follows:
\begin{itemize}
    \item We propose {BEV-Conditioned Controllable Occupancy Generation}, which consists of a BEV-based Occupancy VAE and a Spatio-Temporal Occupancy Diffusion Transformer (STOccDiT). The VAE adopts a compact 2D BEV representation for high-fidelity latent encoding, while STOccDiT jointly tokenizes BEV layouts and occupancy latents to generate occupancy in an autoregressive manner. This module enables precise, high-frequency control from arbitrary, cross-dataset, and user-defined BEV inputs.

    \item We design {Geometry-Grounded View Expansion (GGVE)}, a novel paradigm that treats the generated occupancy sequence as the canonical spatial anchor. By rendering explicit geometric conditioning signals from occupancy, GGVE enables autoregressive, reference-free multi-view video synthesis with arbitrary camera counts and poses, overcoming the reference-frame dependency that limits existing occupancy-guided methods.

    \item We curate {nuCraftv2}, a $12$\,Hz high-frequency benchmark with synchronized BEV layouts and high-quality dense semantic occupancy. Built upon this benchmark, \MethodName achieves state-of-the-art performance in both occupancy and video generation, and demonstrates tangible benefits for applications.
\end{itemize}
\section{Related Works}
\label{sec:related_works}

\subsection{Controllable Occupancy Generation}

Semantic occupancy has emerged as a structured volumetric representation that encodes both geometry and semantics, making it suitable for controllable scene generation. 
Early efforts on layout-conditioned occupancy synthesis include UrbanDiffusion~\cite{zhang2024urban}, which generates semantic occupancy from BEV maps using a 3D diffusion model. 
More recent works have further explored controllable generation from BEV or layout conditions, such as UniScene~\cite{uniscene}, which proposes an occupancy-centric pipeline for generating semantic occupancy from BEV layouts; InfiniCube~\cite{infinicube}, which enables unbounded and controllable dynamic 3D scene generation; WoVoGen~\cite{wovogen}, which develops a world volume-aware diffusion model for controllable multi-camera driving scene generation; X-Scene~\cite{x-scene}, which advances large-scale generation with flexible multi-granular control; and ConsistentCity~\cite{zhu2025consistentcity}, which improves temporal consistency under BEV conditioning via a semantic flow-guided Occupancy DiT.

However, most existing methods still struggle to achieve strong controllability from arbitrary or user-defined BEV inputs. 
Our work addresses this limitation by developing an autoregressive occupancy generator that enables precise and high-frequency control from flexible BEV layouts.


\subsection{Controllable Video Generation}

The advent of general-purpose video diffusion models~\cite{blattmann2023align,blattmann2023stable} has spurred significant progress in driving-specific video generation.
Early single-view world models~\cite{drivegan,drivedreamer,genad} were constrained to short sequences, whereas more recent methods scale to longer, higher-resolution videos with enhanced trajectory and action conditioning~\cite{gao2024vista,driving_world,epona,hu2023gaia,gaia-2}.
A parallel line of research targets \emph{multi-view} driving video generation conditioned on BEV layouts, HD maps, or 3D bounding boxes, exemplified by MagicDrive~\cite{gao2024magicdrive} and its long-video extension MagicDrive-V2~\cite{gao2025magicdrivev2}, as well as Panacea~\cite{wen2024panacea}, DriveDreamer-2~\cite{drivedreamer-2}, SubjectDrive~\cite{SubjectDrive}, UniMLVG~\cite{UniMLVG}, CogDriving~\cite{CogDriving}, DiVE~\cite{DiVE}, and Glad~\cite{Glad}.
To improve cross-view coherence, existing methods commonly employ lightweight cross-view attention mechanisms for information exchange among camera views. Occupancy-guided pipelines~\cite{wovogen,uniscene,infinicube} further introduce 3D structural priors by rendering occupancy representations into semantic or depth maps as conditioning signals for video generation. However, these methods still rely on indirect structural guidance or limited cross-view interactions, which makes it challenging to enforce precise geometric correspondence across cameras. In addition, jointly generating multiple views remains computationally expensive, limiting their efficiency and scalability.

\subsection{Generative Simulation for Autonomous Driving}

Scalable and realistic simulation is fundamental to the development and validation of autonomous driving systems. 
Traditional reconstruction-based simulators, which rely on neural radiance fields~\cite{yang2023unisim} or 3D Gaussian Splatting~\cite{yan2024street,chen2025omnire}, can achieve high visual fidelity when replaying logged trajectories. However, their per-scene optimization inherently limits scenario diversity.

To overcome this limitation, recent research has increasingly focused on generative approaches capable of synthesizing novel driving scenarios. 
Early generative simulators primarily targeted video synthesis, such as DriveDreamer-2~\cite{drivedreamer-2} and the GAIA series~\cite{hu2023gaia,gaia-2}, which employ video diffusion models to generate long-horizon driving videos under trajectory or action conditioning. 
To improve geometric consistency and multi-modal alignment, more recent works have explored unified frameworks that jointly generate multiple data modalities. 
UniScene~\cite{uniscene} introduces an occupancy-centric pipeline that first generates semantic occupancy from BEV layouts and subsequently produces aligned multi-view videos and LiDAR. 
X-Scene~\cite{x-scene} and Cosmos-Drive-Dreams~\cite{ren2025cosmos} further advance large-scale multi-modal generation by improving spatio-temporal consistency. 
DriveArena~\cite{yang2025drivearena} and ReSim~\cite{yang2026resim} explore controllable and large-scale generative simulation using autoregressive models and heterogeneous training data.

These works reflect a growing trend toward controllable and multi-modal generative simulation. 
In contrast to these prior efforts, our framework positions forecasted occupancy as the central representation and integrates autoregressive occupancy generation with geometry-grounded view expansion to enable high-fidelity and flexible scene synthesis from arbitrary BEV layouts.
\section{Method}
\label{sec:method}

\begin{figure*}[t]
  \centering
   \includegraphics[width=1.0\linewidth]{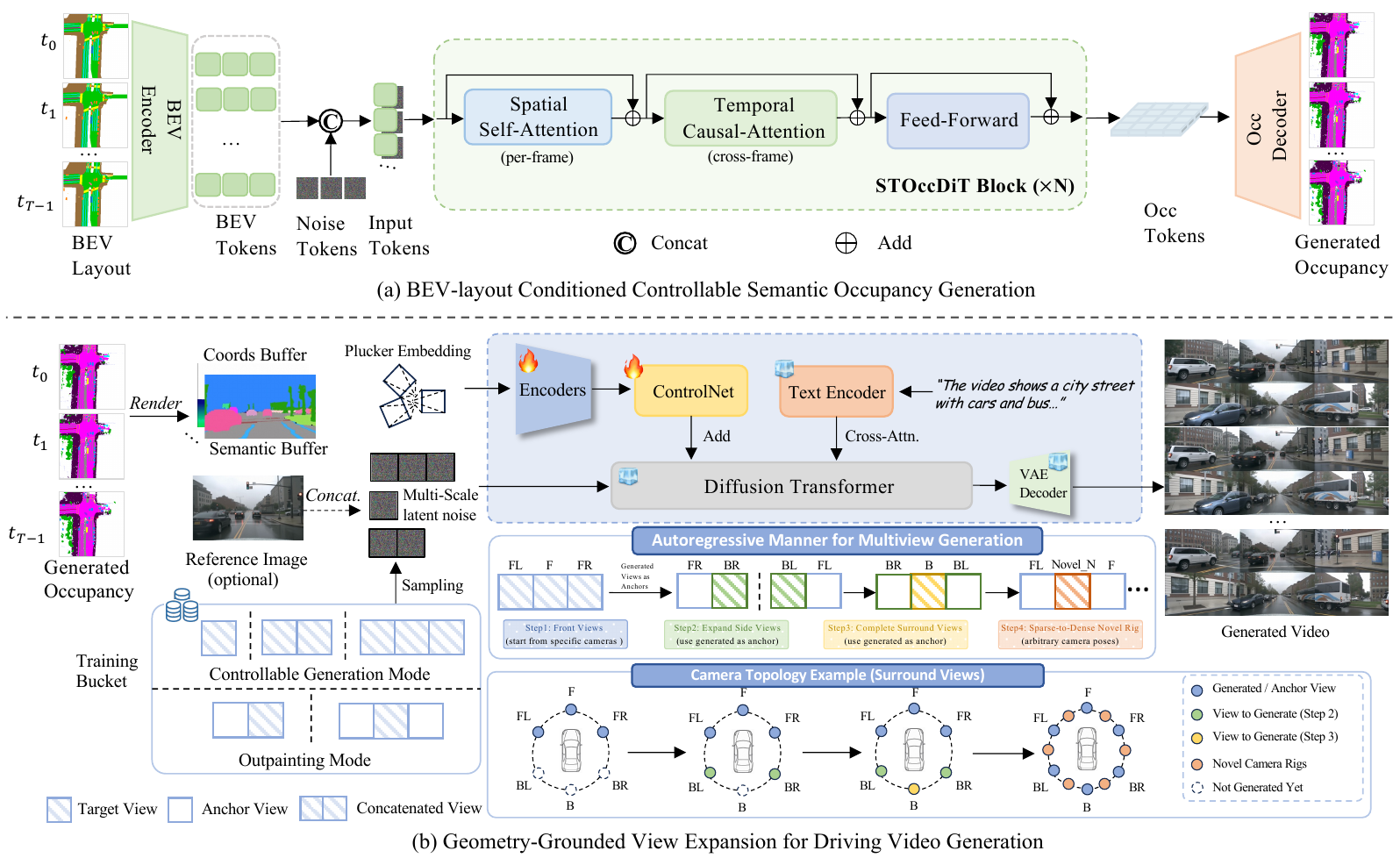}

   \caption{\textbf{Overall framework of \MethodName.}
    \textbf{(a) BEV-layout conditioned controllable semantic occupancy generation}. BEV layout sequences serve as conditions and are fed into the spatial-temporal occupancy diffusion transformer to generate corresponding semantic occupancy sequences. 
    \textbf{(b) Geometry-grounded view expansion for driving video generation.} The generated occupancy can be rendered into coordinate and semantic buffers, which, together with the Pl\"ucker embeddings of the camera poses, are encoded as conditioning inputs to the ControlNet. Multi-view video generation is performed in an autoregressive manner by sampling training examples with varying view combinations from a training bucket. This design enables the synthesis of novel views and supports an arbitrary number of camera views.
    }
   \label{fig:pipeline}
   \vspace{-1.2em}
\end{figure*}




\subsection{Problem Formulation}
\label{sec:method:problem}

Given a sequence of BEV semantic layouts \(\mathcal{B}_{1:T}\), our goal is to generate a temporally consistent 3D semantic occupancy sequence \(\mathcal{O}_{1:T}\) and the corresponding multi-view driving video sequence \(\mathbf{V}_{1:T,1:K}\). 
We formulate the generation process as the composition of two stages with semantic occupancy as the central representation:
\begin{equation}
p(\mathbf{V}_{1:T,1:K} \mid \mathcal{B}_{1:T})
=
p(\mathbf{V}_{1:T,1:K} \mid \mathcal{O}_{1:T})
\cdot
p(\mathcal{O}_{1:T} \mid \mathcal{B}_{1:T}).
\end{equation}

In the first stage, we generate the occupancy sequence from the input BEV layouts using a BEV-conditioned generative model. 
This model produces high-fidelity, temporally consistent occupancy that supports precise control from arbitrary, cross-dataset, and user-defined BEV inputs.
In the second stage, we synthesize multi-view driving videos directly from the generated occupancy sequence. 
A Geometry-Grounded View Expansion module takes \(\mathcal{O}_{1:T}\) as input and produces a temporally and cross-view consistent video sequence \(\mathbf{V}_{1:T,1:K}\), where both the number of camera views \(K\) and the camera poses can be flexibly specified at inference time.

This occupancy-centric formulation allows the entire framework to be evaluated using only BEV layouts as input, while enabling high-fidelity occupancy generation and flexible multi-view video synthesis in a unified pipeline.

\begin{figure*}[t]
    \centering
    \includegraphics[width=1.0\linewidth]{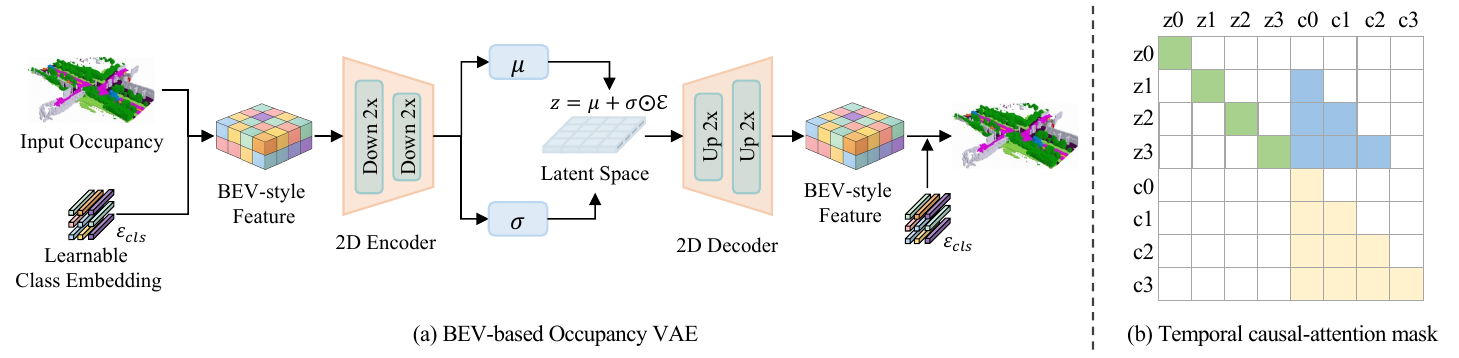}
    \caption{(a) \textbf{BEV-based occupancy VAE} with 2D encoder and decoder. (b) The customized \textbf{temporal causal-attention mask} used in the STOccDiT block.}
    \label{fig:vae_and_mask}
    \vspace{-1.2em}
\end{figure*}

\subsection{Controllable Semantic Occupancy Generation}
\label{sec:method:occ}

\myparagraph{BEV-based Occupancy VAE.}
To obtain a compact yet expressive representation for occupancy, we employ a fully 2D variational autoencoder. The overview of the VAE is illustrated in Fig.~\ref{fig:vae_and_mask} (a).

Specifically, given a 3D occupancy representation \( \mathcal{O}_t \in \mathbb{R}^{H \times W \times Z} \), we first convert it into a 2D BEV representation \( \tilde{\mathcal{O}}_t \in \mathbb{R}^{H \times W \times Z C'} \) following~\cite{zheng2023occworld}. Concretely, each occupancy label are mapped to a learnable \( C' \)-dimensional class embedding, and the resulting height-wise embeddings are then concatenated along the channel dimension at each BEV location. 
We then encode \( \tilde{\mathcal{O}}_t \) into a compact latent feature \( \mathbf{z}_t \in \mathbb{R}^{h \times w \times C_z} \) using a 2D encoder built with 2D convolutional layers and 2D axial attention layers. Here, \( h \) and \( w \) denote the downsampled spatial dimensions, and \( C_z \) denotes the latent channel dimension. 
The 2D decoder reconstructs the resulting latent into an occupancy representation \( \widehat{\mathcal{O}}_t \in \mathbb{R}^{H \times W \times Z} \) using a symmetric 2D decoder composed of the same building blocks and . 
This fully 2D design can significantly improve computational efficiency over volumetric and triplane architectures~\cite{zhang2024urban,dynamiccity,geniedrive} while preserving the spatial details, yielding a compatible 2D latent space for most of the 2D image or video diffusion models~\cite{wan2025wan2.1}.

Previous occupancy VAE~\cite{gu2024dome, zheng2023occworld} typically optimize the standard cross-entropy loss. We instead adopt focal loss~\cite{Linfocalloss} as the main reconstruction objective, motivated by the severe class imbalance in occupancy prediction. By down-weighting easy samples, focal loss can encourage the model to focus on hard and under-represented regions. We further incorporate the Lovasz-Softmax loss to improve region-level prediction quality and use a KL divergence term to regularize the latent distribution. The overall training objective is
\begin{equation}
\mathcal{L}_{\mathrm{occ}}^{\mathrm{vae}}
=
\mathcal{L}_{\mathrm{focal}}
+
\lambda_{1}\mathcal{L}_{\mathrm{LS}}
+
\lambda_{2}\mathcal{L}_{\mathrm{KL}},
\end{equation}
where \( \mathcal{L}_{\mathrm{focal}} \) denotes the focal loss, \( \mathcal{L}_{\mathrm{LS}} \) denotes the Lovasz-Softmax loss, and \( \mathcal{L}_{\mathrm{KL}} \) is the KL divergence loss. Here, \( \lambda_{1} \) and \( \lambda_{2} \) are balancing weights.
Once trained, the encoder and decoder are frozen, and only the latent $\mathbf{z}_t$ is used for subsequent generative modeling.

\myparagraph{BEV Layout Encoding.}
Most of the previous methods regard the semantic BEV layout as a plain 2D image and using simple convolutional layers to encode it~\cite{uniscene,zhang2024urban} directly, missing the semantic information. 
We design a dedicated BEV layout encoder that respects the multi-hot nature of the input. 
Each semantic class is assigned a learnable embedding vector. 
For every spatial location, the embeddings of active classes are summed, while locations without any active class receive a learnable empty embedding. 
This design provides a clear representation of background regions and avoids ambiguity between absent labels and zero-padding. 
The resulting feature map is downsampled to align with the occupancy latent resolution and flattened into tokens.

\myparagraph{Spatial-Temporal Occupancy DiT.}
As illustrated in Fig.~\ref{fig:pipeline}, we model the conditional latent distribution $p(\mathbf{z}_{1:T} \mid \mathcal{B}_{1:T})$ using a causal spatial-temporal diffusion transformer, which consists of a stack of identical STOccDiT blocks. 
At each timestep, the noisy latent $\mathbf{z}_t^{\tau}$ is concatenated with the BEV layout feature $\mathcal{B}_t$ along the token dimension as the input tokens before being processed by the network.
Each STOccDiT contains a spatial self-attention operation within the current frame to capture spatial dependencies, and a temporal causal attention operation across multiple frames to capture temporal dependencies. After that, a feed-forward operation obtains the outputs.

STOccDiT is optimized using the rectified flow-matching strategy in the latent space. 
In contrast to most existing BEV-conditioned occupancy generators that learn a direct one-to-one mapping, our model is designed to be \emph{autoregressive} across time. 
During training, both the noisy latent of the current frame $z_t$ and the clean latents of previous frames $c_t$ are provided to the model under a causal temporal mask, as depicted in Fig.~\ref{fig:vae_and_mask} (b). 
This dual-stream design, analogous to teacher forcing, enables the model to learn strong conditional control from BEV layouts while allowing fully parallel training. 
At inference, the model generates occupancy autoregressively by conditioning on its own previously generated frames. 
This formulation not only improves fine-grained controllability but also naturally supports long-horizon occupancy sequence generation.

The detailed architecture of the STOccDiT block, flow-matching objective, and causal attention mechanism are described in the supplementary material.

\subsection{Geometry-Grounded View Expansion}
\label{sec:ggve}

Most existing video diffusion models are conditioned on text, single images, or reference clips, which provide limited control over 3D scene geometry and arbitrary camera configurations. 
To overcome this limitation, we introduce Geometry-Grounded View Expansion \textbf{(GGVE)}, a module that generates temporally and cross-view consistent multi-view driving videos directly from generated occupancy sequences, as depicted in Fig.~\ref{fig:pipeline} (b).

GGVE treats the generated volumetric occupancy as the canonical 3D scene representation. 
For each camera view, three geometric signals are rendered from the occupancy volume using $f$VDB~\cite{williams2024fvdb}, following InfiniCube~\cite{infinicube}: a semantic buffer, a coordinate buffer, and the Pl\"ucker embedding~\cite{yang2026neoverse} of the camera pose.
These explicit geometric cues enable the model to perform geometry-driven video synthesis, where the camera rig becomes a flexible, query-time variable.

The GGVE module builds upon the Wan2.1-T2V-14B~\cite{wan2025wan2.1} backbone and adopts a VACE-style~\cite{jiang2025vace} ControlNet architecture, where a trainable control branch runs alongside the frozen DiT backbone to process per-camera geometric conditions, including reference-view RGB latents, Plücker camera embeddings, and semantic/coordinate buffer features. The resulting hint tensors are propagated back into the backbone via uniformly spaced residual connections, preserving the pretrained generative prior while enabling precise geometric control over generation. Text prompts describing the driving environment are encoded as additional conditioning inputs to control scene attributes, and an optional reference camera image can further be provided to enforce appearance consistency. The model is trained with a unified mode-mixture strategy that jointly optimizes multi-view text-to-video and various outpainting configurations, while keeping the DiT backbone, text encoder, and VAE frozen throughout. At inference, GGVE synthesizes surround-view videos \emph{autoregressively} from the occupancy sequence alone: starting from the front cameras, the model progressively expands to the remaining views by leveraging previously generated frames as anchors, a mechanism that naturally supports arbitrary camera rigs by allowing virtual views to be inserted at any desired pose. By grounding video generation in explicit 3D occupancy, GGVE enables flexible, reference-free synthesis of multi-view driving videos with arbitrary camera counts, poses, and temporal horizons within a single unified model.

\section{Experiments}
\label{sec:experiments}

\subsection{Datasets}
\label{sec:exp:data}

\myparagraph{nuCraftv2.}
We train \MethodName on the nuCraftv2 dataset to enable high-frequency temporal modeling and high-resolution occupancy generation. 
NuCraftv2 is a 12\,Hz resampling of the nuScenes~\cite{caesar2020nuscenes} dataset that provides synchronized high-resolution 3D semantic occupancy, HD-map annotations, and 3D bounding boxes at every keyframe. 
Details of the dataset construction are described in the supplementary material.

BEV layouts are constructed from the nuScenes HD-map and 3D bounding box annotations. 
Each layout is rasterized at a resolution of 0.4\,m on a $256 \times 256$ grid with 15 multi-hot channels, representing 10 dynamic agent classes and 5 static map primitives. 
Unless otherwise specified, we adopt an occupancy volume spanning $[-51.2, 51.2]^2 \times [-5.0, 5.0]$\,m around the ego vehicle, with a voxel size of 0.4\,m. 
This yields a tensor of shape $256 \times 256 \times 25$ covering 21 semantic classes, including a distinguished free-space label.
The multi-view camera images are 12Hz and the same with the original nuScenes dataset.

\begin{figure*}[t]
    \centering
    \includegraphics[width=\linewidth]{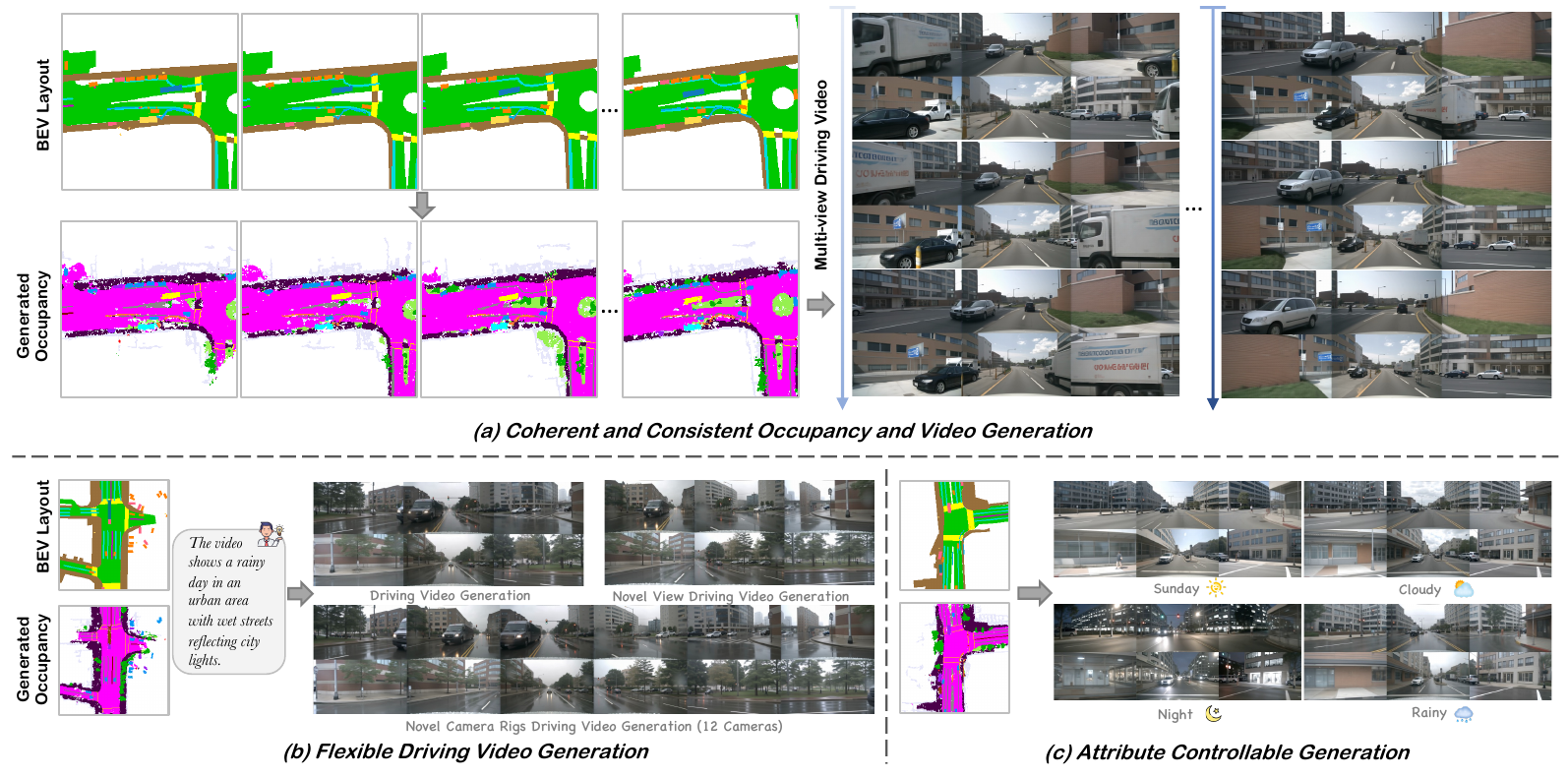}
    \caption{\textbf{Versatile generation ability of \MethodName.} (a) Coherent and temporally consistent occupancy and multi-view video generation conditioned on BEV layout sequences. (b) The flexible driving video generation ability of our method. Including high-quality driving video generation from user prompts and occupancy, the novel view driving video generation and arbitrary camera rigs driving video generation. (c) Controllable generation of attribute-diverse videos by modifying the input text prompts. Please zoom in for better results.
    }
    \vspace{-4pt}
    \label{fig:quali_vis}
\end{figure*}

\begin{table*}[t]
    \centering
    \small
    \setlength{\tabcolsep}{4.5pt}
    \renewcommand{\arraystretch}{1.15}
    \begin{tabular*}{\linewidth}{@{\extracolsep{\fill}} l|cc|cc|cc|cc|cc}
        \toprule
        & \multicolumn{2}{c|}{\textbf{3D Volume} ($\uparrow$)}
        & \multicolumn{2}{c|}{\textbf{BEV Top-down} ($\uparrow$)}
        & \multicolumn{2}{c|}{\textbf{BEV vs.\ Layout} ($\uparrow$)}
        & \multicolumn{2}{c|}{\textbf{Cam Render} ($\downarrow$)}
        & \multicolumn{2}{c}{\textbf{BEV Render} ($\downarrow$)} \\
        Method
        & mIoU & IoU
        & mIoU & IoU
        & mIoU & IoU
        & FID  & KID
        & FID  & KID \\
        \midrule
        \MethodName (ours)
            & \textbf{19.01} & \textbf{15.58}
            & \textbf{35.04} & \textbf{60.53}
            & \textbf{52.45} & \textbf{68.97}
            & \textbf{15.03} & \textbf{13.00}
            & \textbf{54.52} & \textbf{41.30} \\
        \bottomrule
    \end{tabular*}
    \caption{\textbf{Occupancy generation quality on {nuCraftv2-val}.}
    For the FID and KID groups, scores are reported on the depth and semantic
    streams jointly ({ average} across the two channels.}
    \label{tab:occ_gen}
    \vspace{-0.7em}
\end{table*}

\begin{table}[t]
\begin{center}
\footnotesize
\renewcommand\tabcolsep{5pt}
	\centering
   \resizebox{0.99\linewidth}{!}{
\begin{tabular}{l|c|c|c}
\toprule Method  & 
\begin{tabular}[c]{@{}c@{}}Downsampled\\Size\end{tabular} & mIoU $\uparrow$ & IoU $\uparrow$ \\ \midrule
OccLLama (VQVAE)~\cite{wei2024occllama} & $128 \times 50 \times 50$ & {75.2}& 63.8\\
OccWorld (VQVAE)~\cite{zheng2023occworld}  & $128 \times 50 \times 50$ & 66.4 & 62.3 \\
OccSora (VQVAE)~\cite{wang2024occsora} & $512 \times 25 \times 25$ & 27.4& 37.0\\
\midrule
$\mathcal{X}$-Scene (Triplane-VAE)~\cite{x-scene} & $32 \times 64 \times 64$ & 58.05 & 44.96 \\
$\mathcal{X}$-Scene (Triplane-VAE)~\cite{x-scene} & $32 \times 128 \times 128$ & 81.47 & 61.94 \\
\midrule
UniScene (VAE)~\cite{uniscene} & $8 \times 50 \times 50$ & 72.90 & 64.10 \\
\rowcolor{gray!10} \textbf{Ours} (VAE) & $32 \times 64 \times 64$ & 76.02 & 62.30 \\
\rowcolor{gray!10} \textbf{Ours} (VAE) & $16 \times 128 \times 128$ & 85.29 & 73.72 \\
\rowcolor{gray!10} \textbf{Ours} (VAE) & $32 \times 128 \times 128$ & \textbf{90.27} & \textbf{81.25} \\
\bottomrule
\end{tabular}
 }
 \caption{\textbf{Occupancy VAE performances.} 
 The downsampled size is represented as the latent shape \((\mathit{C} \times \mathit{H} \times \mathit{W})\), where \(\mathit{C}\) denotes the number of latent channels, while \(\mathit{H}\) and \(\mathit{W}\) denote the spatial resolution. The best results are highlighted in \textbf{bold}.
 }
\vspace{-3em}
\label{tab_occ_rec}
\end{center}
\end{table}

\subsection{Implementation Details}
\label{sec:exp:impl}

 The STOccDiT is trained using PyTorch on eight NVIDIA A100 GPUs with 80\,GB memory each, based on the mmengine~\cite{mmengine2022} framework. 
The effective batch size is set to 2 per GPU for STOccDiT and 8 single-frame grids per GPU for the VAE.
We train the GGVE model for 5k iterations at resolution $480\times832$ using 49 sampled frames per iter. Optimization is performed with AdamW using a learning rate of $2\times10^{-5}$, together with bf16 mixed precision and gradient checkpointing. Training follows a curriculum over single-view, two-view, and three-view T2V settings, as well as outpainting modes. More implementation details of the Occupancy VAE, STOccDiT, and the GGVE module are provided in the supplementary material.

We evaluate \MethodName across two aspects using a range of metrics: \textbf{1) Occupancy Generation:} We evaluate the reconstruction results of the VAE with IoU and mIoU metrics. For occupancy generation, inspired by~\cite{x-scene}, we not only report the mIoU and IoU metrics under 3D volume, BEV top-down and BEV vs. layout settings, but also the generative quality of rendered images from both camera and BEV views using the 2D metrics Fréchet Inception Distance (FID), Kernel Inception Distance (KID). More details about the evaluation metrics are provided in the supplementary.
\textbf{2) Video Generation:} We evaluate video generation quality by using the FVD metric and the controllability by performing BEV segmentation and 3D object detection tasks on the generated data from the nuScenes~\cite{caesar2020nuscenes} validation set.

\subsection{Main Results}

\myparagraph{Versatile Generation Ability.} 
As shown in Fig.~\ref{fig:quali_vis}, \MethodName supports versatile and controllable driving scene generation.
Given an arbitrary BEV layout sequence, \MethodName first generates a temporally coherent semantic occupancy sequence via STOccDiT, which then conditions the GGVE module to produce multi-view videos. 
The resulting occupancy and videos share the same geometric structure, ensuring consistent object identity, scale, and motion across modalities and viewpoints (Fig.~\ref{fig:quali_vis} (a)). 
Beyond the standard camera rig, \MethodName enables novel-view synthesis and arbitrary camera configurations at inference time without retraining (Fig.~\ref{fig:quali_vis} (b)). 
\MethodName also supports fine-grained attribute control through user prompts. 
Specifying conditions such as weather or lighting in the text prompt (e.g., ``cloudy'', ``nighttime'') produces videos that faithfully reflect the desired attributes while preserving the geometry defined by occupancy.
These capabilities highlight \MethodName as a flexible, high-quality generator suitable for diverse simulation and data synthesis tasks.

\begin{figure}[t]
    \centering
    \includegraphics[width=\linewidth]{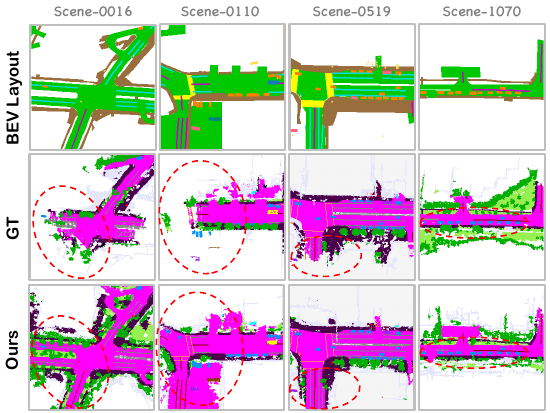}
    \caption{\textbf{Qualitative results of occupancy generation.} Our method could generate faithful and high-fidelity occupancy based on BEV layout as input, even better than GT.
    }
    \label{fig:occ_gen_vis}
    \vspace{-1.2em}
\end{figure}

\begin{figure}[t]
    \centering
    \includegraphics[width=\linewidth]{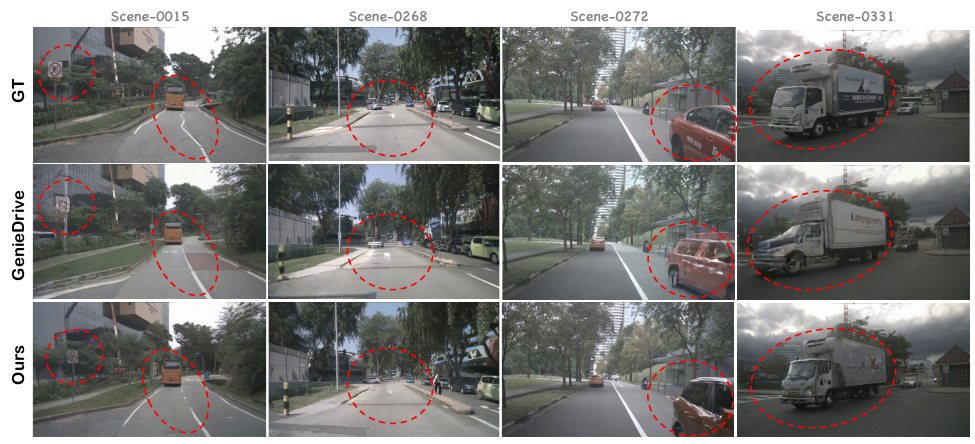}
    \caption{\textbf{Qualitative results of driving video generation.} Our method excels in object structure quality and appearance details.
    }
    \label{fig:video_gen_vis}
    \vspace{-1.2em}
\end{figure}

\myparagraph{Occupancy VAE and Occupancy Generation.}
\label{exp:vae}
Since precisely reconstructing the occupancy is vital for occupancy generation, we first compare our BEV-based occupancy VAE with the existing methods on reconstruction accuracy in Tab.~\ref{tab_occ_rec}. The results demonstrate that our method achieves superior reconstruction performance and consistently outperforms prior approaches. In particular, under the same compression setting (e.g., \(32 \times 128 \times 128\)), it surpasses $\mathcal{X}$-Scene~\cite{x-scene} by 8.8\% in mIoU and 19.31\% in IoU.

The occupancy generation metrics are reported in Tab.~\ref{tab:video_gen_comparison}. For the 3D volume metric, \MethodName achieves \(19.01/15.58\) mIoU/IoU, demonstrating that the model recovers a substantial portion of scene geometry from BEV layout alone. 
Projecting to the BEV plane further improves performance to \(35.04/60.53\) mIoU/IoU, indicating that the model more accurately predicts object locations than their precise vertical extent.
Rendered images from the camera view yield substantially lower FID/KID (\(15.03/13.00\)) than those from the BEV view (\(54.52/41.30\)), as perspective projection is less sensitive to fine-grained height errors. 
The highest IoU is observed in the BEV-versus-Layout setting (\(52.45/68.97\)), which measures how faithfully the model respects the input conditioning rather than direct comparison with ground-truth occupancy. 
This high score confirms that the BEV layout condition is effectively utilized during generation.
We further visualize the generated occupancy in Fig.~\ref{fig:occ_gen_vis}. 
The results demonstrate that our method produces high-fidelity occupancy that faithfully respects the conditioning BEV layout. 
Notably, even when the ground-truth occupancy is incomplete owing to limited LiDAR coverage, our model is still able to generate complete and coherent scene structures.


\myparagraph{Video Generation Results.}
\label{exp:vis}
Because existing baselines are not designed for the view-expansion setting, they cannot be directly evaluated under the full task protocol of GGVE. To enable fair and meaningful comparison, we follow the image-to-video evaluation protocol widely adopted by prior driving video generation methods, in which all approaches are assessed under aligned conditioning and generation configurations. This allows us to isolate and evaluate the controllability of our geometry-grounded conditioning design independently of task-specific factors.

We adopt the evaluation toolboxes from the CODA~\cite{chen2025coda} competition and MagicDriveV2~\cite{gao2025magicdrivev2} to assess both video generation quality and control accuracy. Specifically, video quality is measured by FVD~\cite{unterthiner2019fvd}, while controllability is evaluated using mIoU and mAP. Following prior practice, mIoU and mAP are computed by applying BEVFormer~\cite{li2024bevformer} to the generated videos and comparing the resulting predictions with the corresponding ground-truth annotations.

Quantitative results are reported in Tab.~\ref{tab:video_gen_comparison}. Compared with recent state-of-the-art driving video generation methods, our approach achieves the best FVD across all evaluated frame lengths. In particular, our method obtains FVD scores of $35.98$, $43.19$, and $57.57$ for 8, 16, and 49 frames, respectively, substantially outperforming prior methods such as MagicDriveV2~\cite{gao2025magicdrivev2} and GenieDrive~\cite{geniedrive}. Moreover, our method achieves $38.26$ mIoU and $27.73$ mAP, outperforming GenieDrive by $7.29$ mIoU and $8.56$ mAP. These results demonstrate that our geometry-grounded conditioning not only improves temporal consistency, but also provides stronger semantic and spatial controllability.

We further provide qualitative visualizations of the generated videos in Fig.~\ref{fig:video_gen_vis}. The results show that GGVE can generate temporally coherent and cross-view consistent multi-view driving videos. In addition, GGVE supports flexible camera configurations and allows dynamic camera adjustment at inference time, which is a key advantage over existing methods that rely on fixed conditioning inputs.

Additionally, we provide more qualitative visualization results in supplementary to show more capbilities such as: novel trajectory rendering, denser scene reconstructions, flexiable scene editing, etc.

\begin{table}[t]
\centering
\footnotesize
\setlength{\tabcolsep}{4pt}
\resizebox{0.99\columnwidth}{!}{%
\begin{tabular}{l|c|c|ccc}
\toprule
Method & Frames & Cond. & FVD $\downarrow$ & mIoU $\uparrow$ & mAP $\uparrow$ \\
\midrule
Panacea~\cite{wen2024panacea} & 8 & BEV & 139.00 & - & - \\
Vista~\cite{gao2024vista} & 25 & Action & 112.65 & - & - \\
MagicDrive~\cite{gao2024magicdrive} & 60 & BEV \& 3D Box & 217.94 & 18.27 & 11.86 \\
MagicDrive-V2~\cite{gao2025magicdrivev2} & 16 & BEV \& 3D Box & 94.84 & 20.40 & 18.17 \\
\midrule
WoVoGen~\cite{wovogen} & 6 & Occ & 417.70 & - & - \\
UniScene~\cite{uniscene} & 8 & Occ & 70.52 & 21.75 & 10.32 \\
UniScene-Rollout & 32 & Occ & 610.15 & 18.69 & 6.24 \\
\midrule
\multirow{3}{*}{{GenieDrive}~\cite{geniedrive}} & 8 & \multirow{3}{*}{Occ} & 55.93 & \multirow{3}{*}{30.97} & \multirow{3}{*}{19.17} \\
 & 16 & & 63.65 & & \\
 & 49 & & 92.52 & & \\
\midrule
\rowcolor{gray!10} & 8 & & \textbf{35.98} & & \\
\rowcolor{gray!10} & 16 & & \textbf{43.19} & & \\
\rowcolor{gray!10} \multirow{-3}{*}{\textbf{Ours}} & 49 & \multirow{-3}{*}{Occ} & \textbf{57.57} & \multirow{-3}{*}{\textbf{38.26}} & \multirow{-3}{*}{\textbf{27.73}} \\
\bottomrule
\end{tabular}%
}
\caption{\textbf{Comparison of video generation methods} on FVD, mIoU, and mAP metrics.}
\label{tab:video_gen_comparison}
\vspace{-1.0em}
\end{table}

\subsection{Ablation Studies}

\myparagraph{Effect of Designs in Occupancy Generation Model.}
We conduct ablation studies on three key design choices in STOccDiT, as reported in Tab.~\ref{tab:ablation_stoccdit}.
Removing the teacher-forcing strategy causes a severe performance drop, with mIoU decreasing from \(19.61\) to \(3.23\) and geometric IoU from \(11.94\) to \(0.98\). Without access to clean history tokens, the model must jointly learn denoising and temporal dynamics from the same gradients, leading to rapid error accumulation during autoregressive rollout.
Reducing the backbone depth from \(L=24\) to \(L=16\) blocks results in a moderate decline of \(4.25\) in mIoU, while geometric IoU remains largely unaffected. This indicates that additional capacity primarily improves semantic classification of occupied voxels rather than basic occupancy detection.
Replacing token concatenation with element-wise addition for BEV conditioning degrades mIoU by \(6.02\) and IoU by \(4.34\). Token concatenation allows direct interaction between BEV layout and occupancy tokens via self-attention, enabling fine-grained spatial alignment that additive fusion cannot achieve.

\myparagraph{Effect of Designs in GGVE Module.}
We ablate the two conditioning signals and report FVD using 16 frames in Tab.~\ref{tab:ablation_fvd}. The full model achieves an FVD of 43.19. Removing either conditioning signal leads to a consistent degradation in generation quality, while removing both results in a strict and substantial performance drop, confirming that each signal contributes non-trivially to the overall generation quality.

\begin{table}[t]
    \centering
    \small
    \setlength{\tabcolsep}{10pt}
    \renewcommand{\arraystretch}{1.1}
    
    \begin{tabular}{l|cc}
        \toprule
        \textbf{Variant} & mIoU\,$\uparrow$ & IoU\,$\uparrow$ \\
        \midrule
        \rowcolor{gray!10}
        \textbf{Ours} (TF, $L\!=\!24$, token concat)  & \textbf{19.61} & \textbf{11.94} \\
        \midrule
        w/o.\ Teacher-forcing                          &  3.23 &  0.98 \\
        w/o.\ Depth $L\!=\!24$ (use $L\!=\!16$)       & 12.22 & 10.11 \\
        Additive BEV (w/o.\ token concat)             & 13.59 &  7.60 \\
        \bottomrule
    \end{tabular}
    \caption{\textbf{Ablation} for designs in the occupancy generation model.
    }
    \label{tab:ablation_stoccdit}
    \vspace{-0.5em}
\end{table}

\begin{table}[t]
\centering
\renewcommand{\arraystretch}{1.1}
\small
\begin{tabular*}{\columnwidth}{@{\extracolsep{\fill}} l|cc}
\toprule
\textbf{Variant} & Removed Condition & FVD $\downarrow$ \\
\midrule
\textbf{Ours}            & ---     & \textbf{43.19} \\
\midrule
\text{w/o Semantic}    & Semantic buffer     & 47.05  \\
\text{w/o Coordinate}  & Coordinate buffer   & 52.59  \\
\text{w/o Both}      & All buffer & 112.20 \\
\bottomrule
\end{tabular*}
\caption{\textbf{Ablation on conditioning signals} of controllable generation.}
\label{tab:ablation_fvd}
\end{table}



\section{Conclusion and Future Work}

In this paper, we presented \MethodName, a unified occupancy-centric framework for controllable driving scene generation. 
\MethodName consists of a BEV-based Occupancy VAE, a Spatio-Temporal Occupancy Diffusion Transformer (STOccDiT), and a Geometry-Grounded View Expansion (GGVE) module. 
Together, these components enable precise high-frequency control from BEV layouts to occupancy and support reference-free multi-view video generation with arbitrary camera configurations.
Extensive experiments show that \MethodName achieves state-of-the-art performance in both occupancy and video generation. 
It also generalizes well to cross-dataset and user-defined BEV layouts and provides measurable benefits for downstream tasks such as sparse-view 3D reconstruction.

\myparagraph{Limitations and Future Work.}
The current framework does not model traffic flow or reactive agent behavior; all agents are conditioned on fixed BEV layouts and do not interact with each other or with the ego vehicle. 
A promising direction is to integrate \MethodName with a traffic behavior model to enable reactive and interactive agent synthesis. 
Furthermore, incorporating \MethodName into a closed-loop simulation platform would allow systematic generation of targeted corner cases and long-tail scenarios, facilitating both training and rigorous evaluation of autonomous driving algorithms under diverse and controllable conditions.




{
    \small
    \bibliographystyle{ieeenat_fullname}
    \bibliography{main}
}

\clearpage
\label{sec:suppl}
\renewcommand{\thesection}{\Alph{section}}
\renewcommand{\thesubsection}{\thesection.\arabic{subsection}}
\setcounter{page}{1}
\setcounter{section}{0}
\maketitlesupplementary

\section{nuCraftv2 Data Curation}

\begin{figure*}[t]
    \centering
    \includegraphics[width=\textwidth]{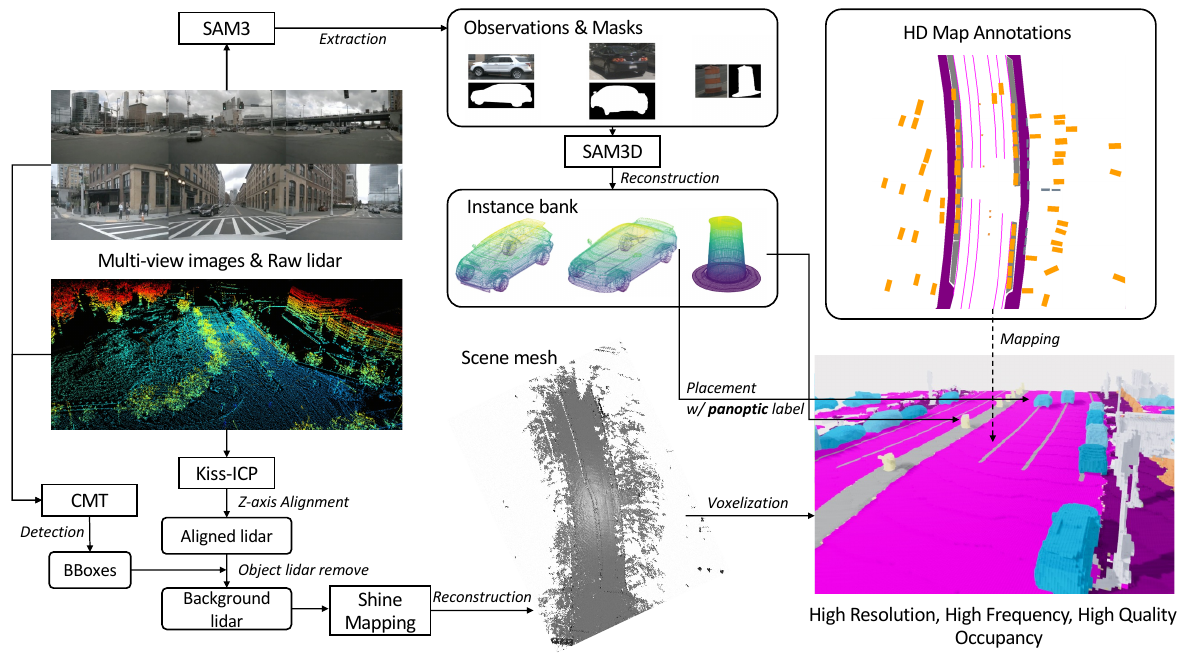}
    \caption{The curation pipeline of nuCraftv2 dataset.}
    \label{Fig: nucraftv2_pipeline}
\end{figure*}

\begin{figure*}[t]
    \centering
    \includegraphics[width=\textwidth]{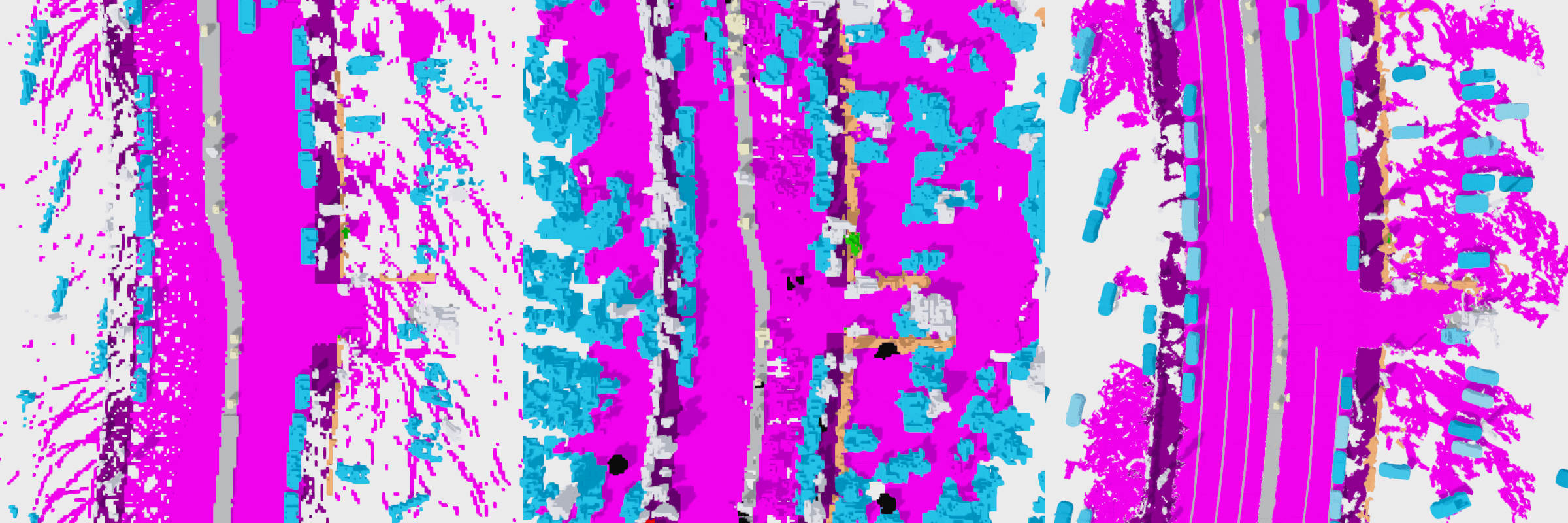}\\[2pt]
    \includegraphics[width=\textwidth]{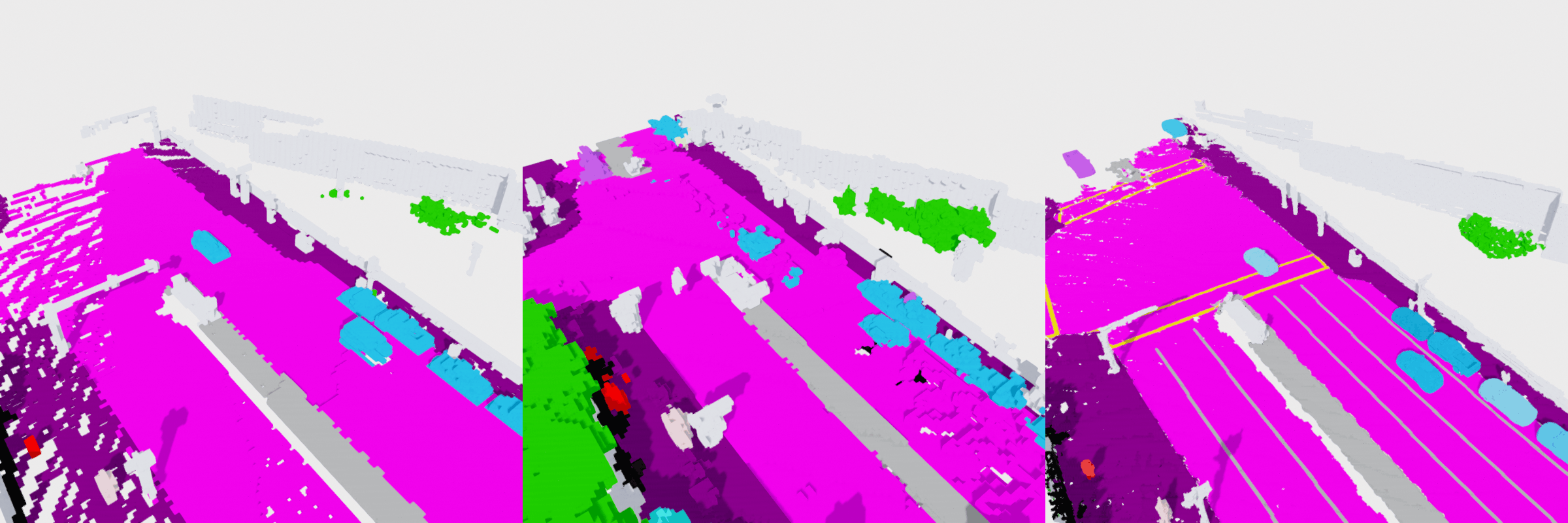}\\[2pt]
    \includegraphics[width=\textwidth]{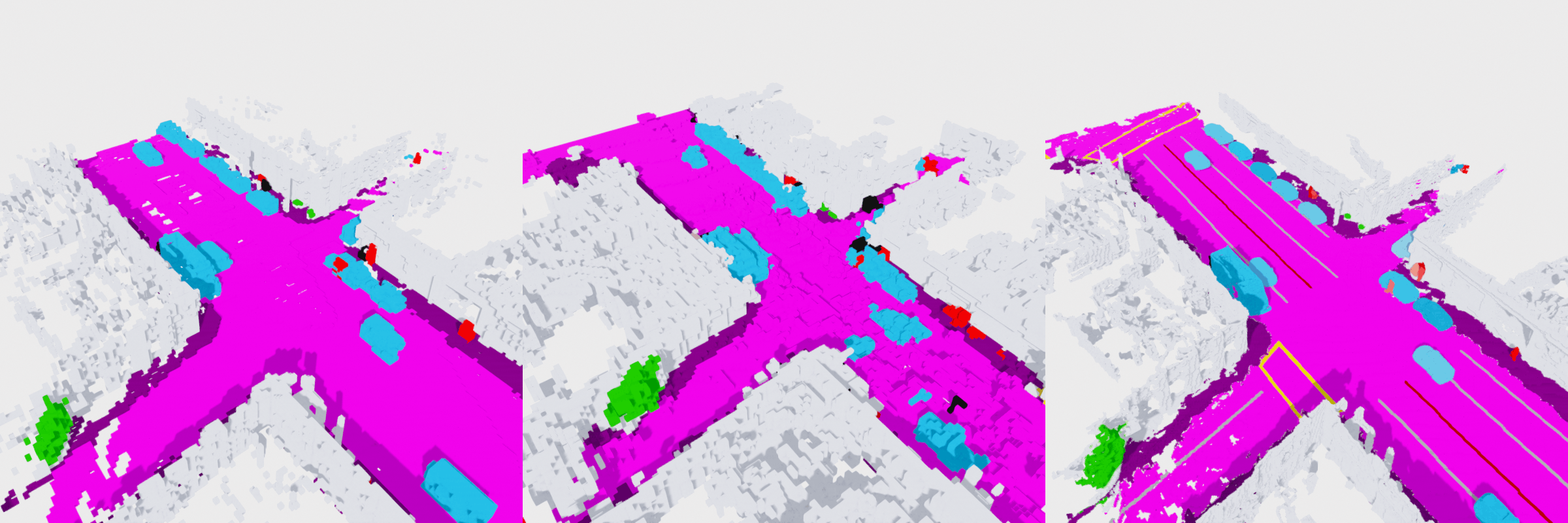}\\[2pt]
    \caption{Qualitative comparison of occupancy visualizations. From left to right: OccWorld  ($2\,\mathrm{Hz}$) \cite{zheng2023occworld}, UniScene ($12\,\mathrm{Hz}$) \cite{uniscene}, and nuCraftv2 ($20\,\mathrm{Hz}$) (ours).
}
    \label{fig:occ_compare_1}
\end{figure*}

\begin{figure*}[t]
    \centering
    \includegraphics[width=\textwidth]{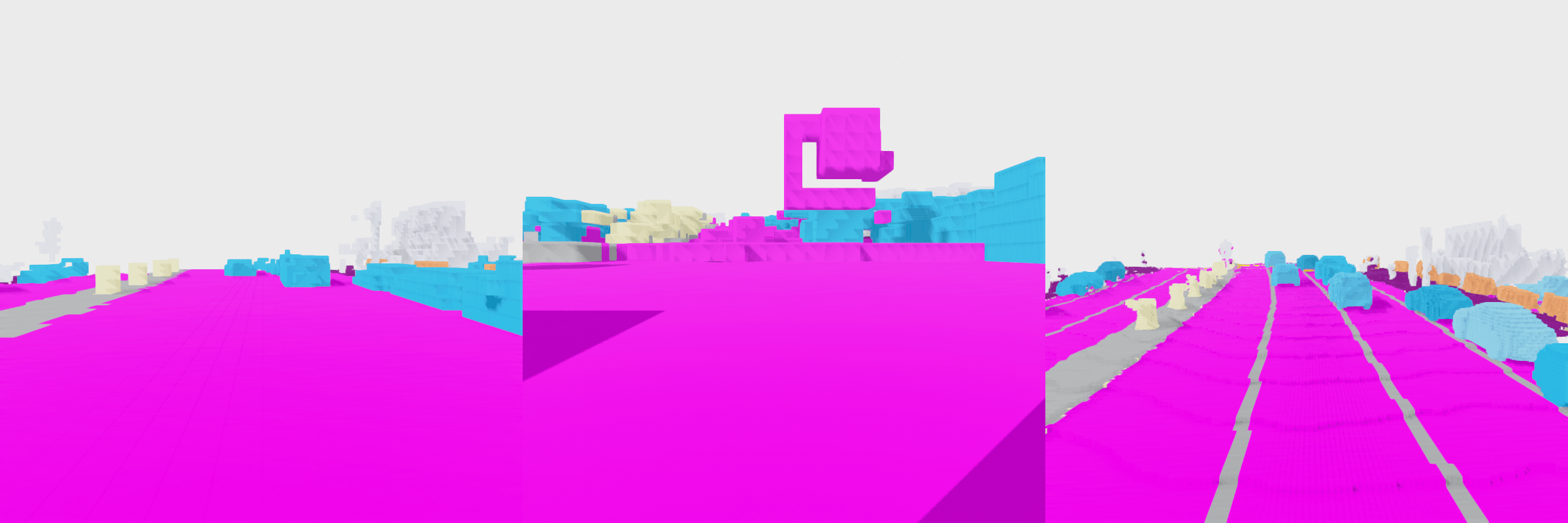}\\[2pt]
    \includegraphics[width=\textwidth]{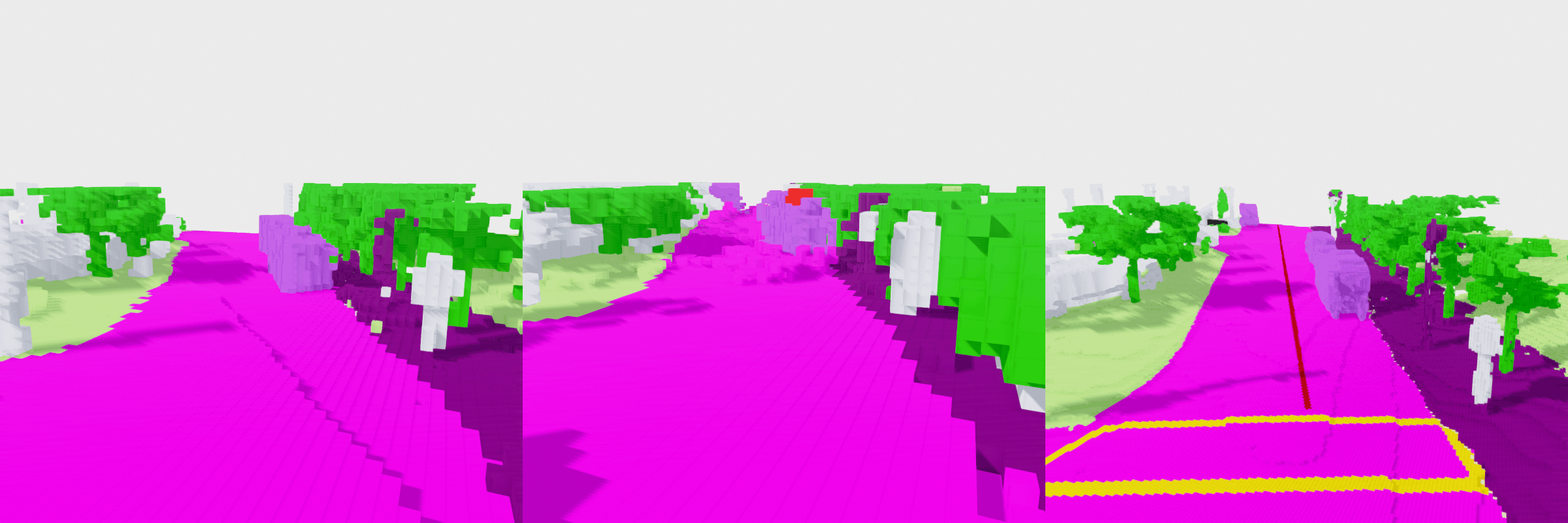}\\[2pt]
    \includegraphics[width=\textwidth]{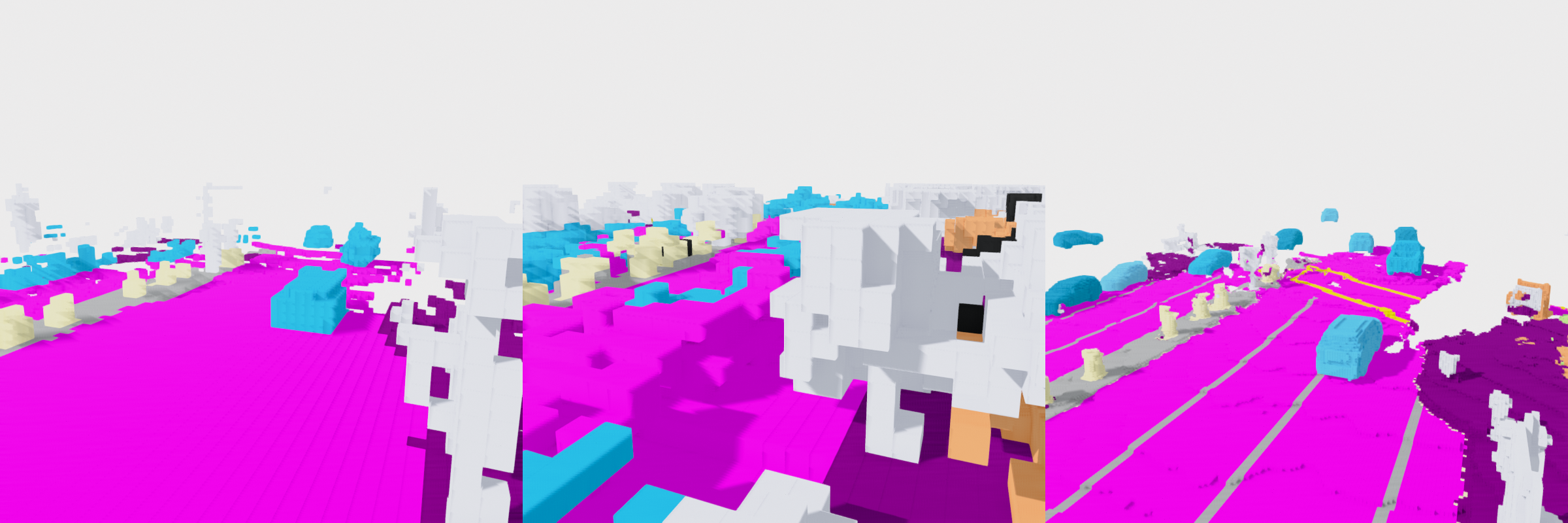}
    \caption{Additional qualitative comparison of occupancy visualizations. From left to right: OccWorld ($2\,\mathrm{Hz}$) \cite{zheng2023occworld}, UniScene ($12\,\mathrm{Hz}$) \cite{uniscene}, and nuCraftv2 ($20\,\mathrm{Hz}$) (ours).}
    \label{fig:occ_compare_2}
\end{figure*}

\begin{figure*}[t]
    \centering

    \begin{minipage}{0.49\textwidth}
        \centering
        \includegraphics[width=\linewidth]{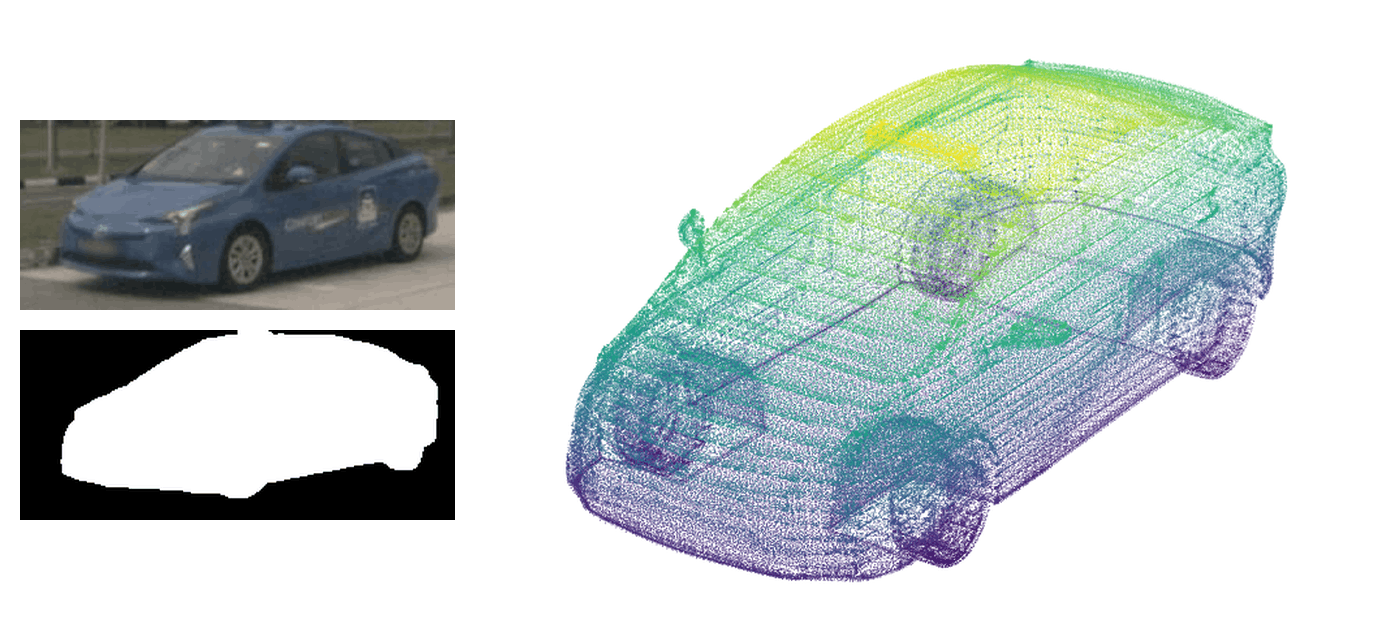}
    \end{minipage}
    \hfill
    \begin{minipage}{0.49\textwidth}
        \centering
        \includegraphics[width=\linewidth]{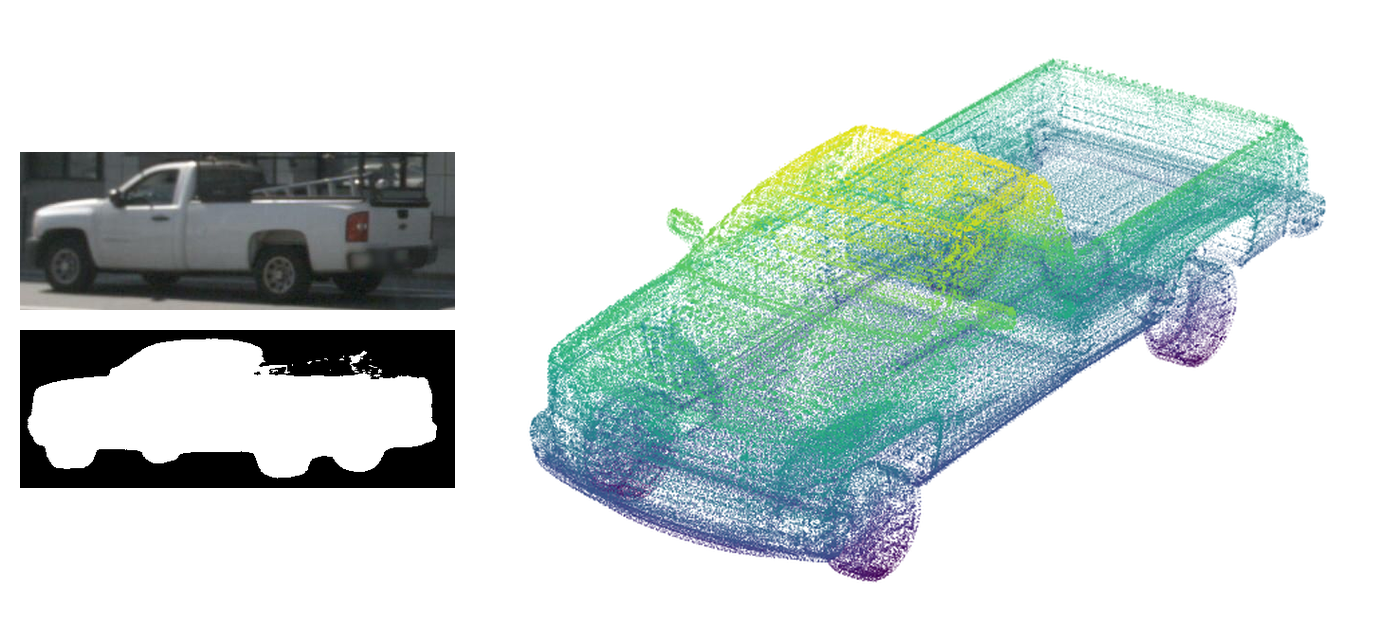}
    \end{minipage}

    \vspace{2pt}

    \begin{minipage}{0.49\textwidth}
        \centering
        \includegraphics[width=\linewidth]{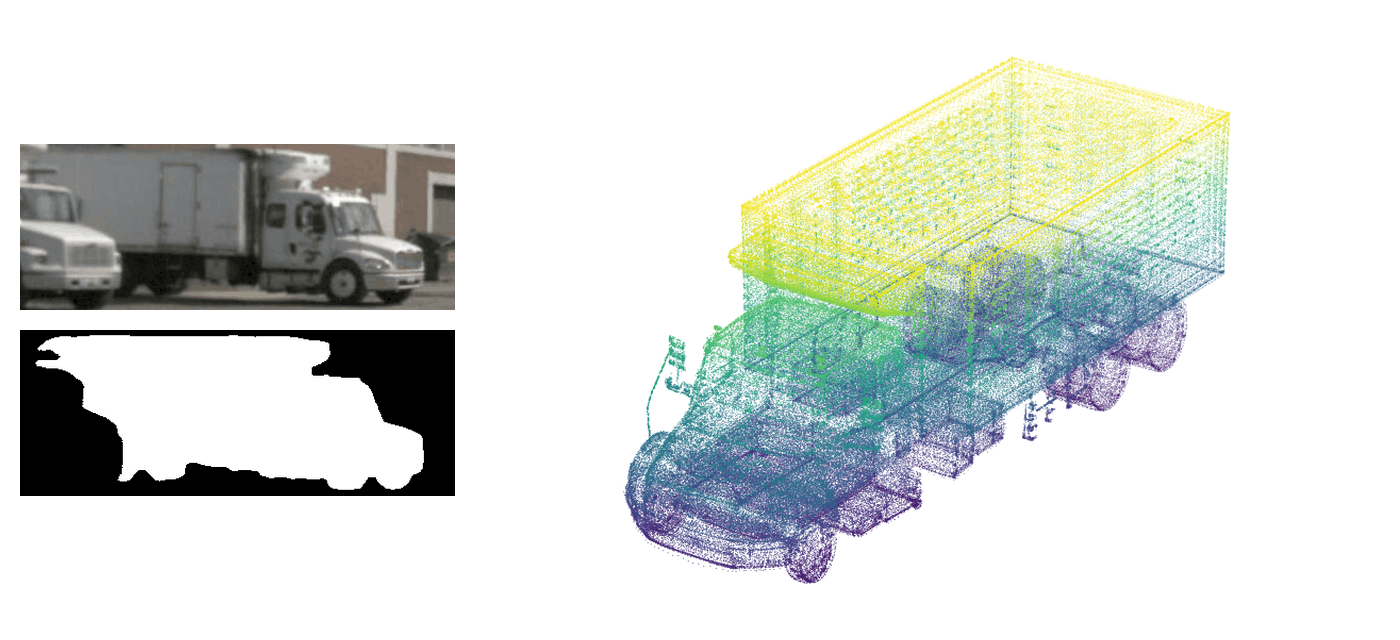}
    \end{minipage}
    \hfill
    \begin{minipage}{0.49\textwidth}
        \centering
        \includegraphics[width=\linewidth]{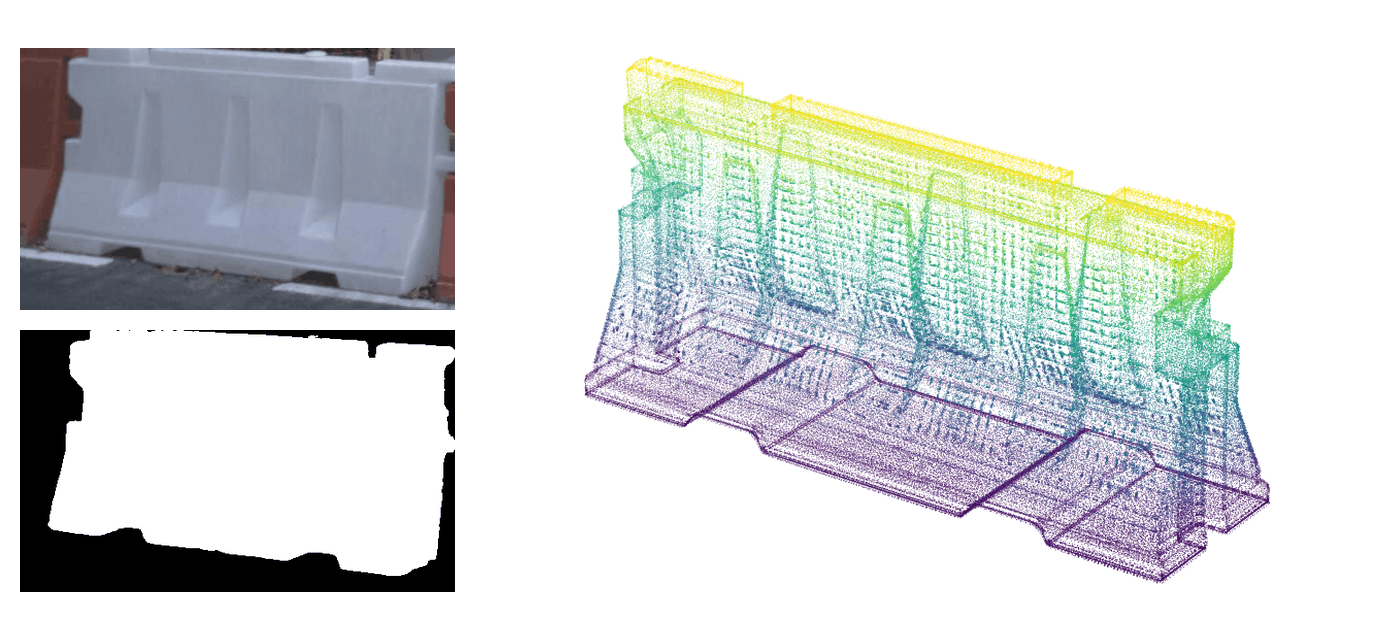}
    \end{minipage}

    \vspace{2pt}

    \begin{minipage}{0.49\textwidth}
        \centering
        \includegraphics[width=\linewidth]{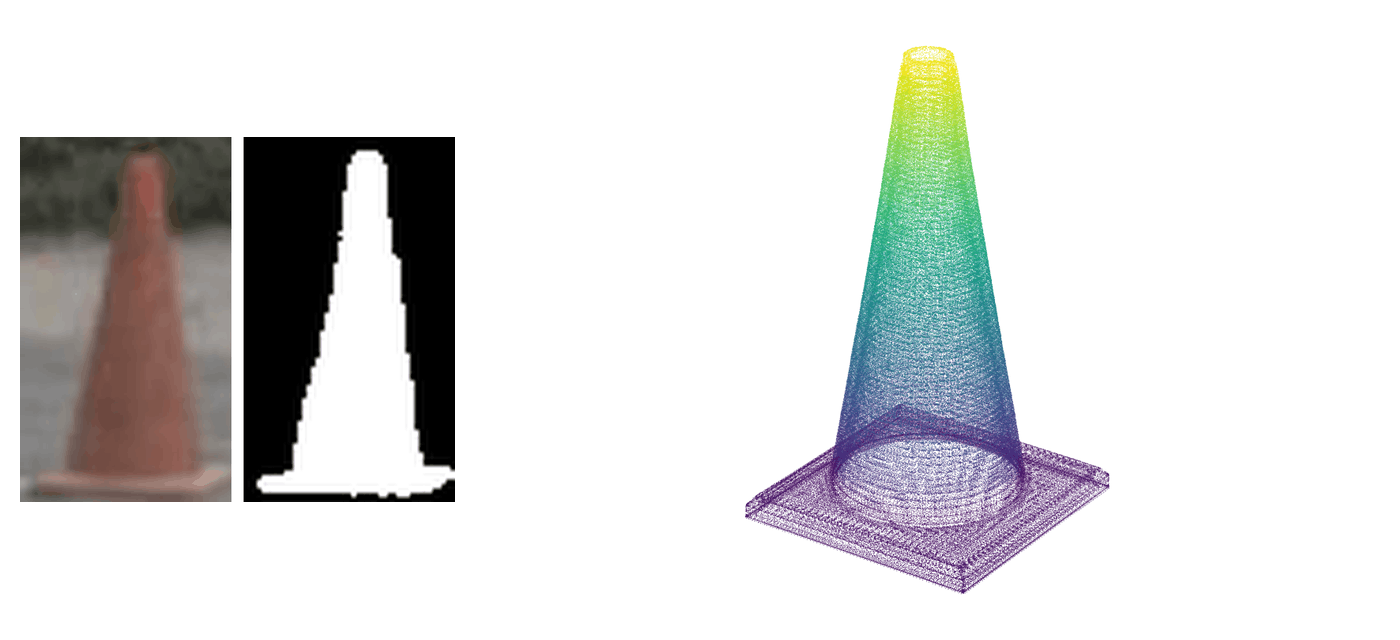}
    \end{minipage}
    \hfill
    \begin{minipage}{0.49\textwidth}
        \centering
        \includegraphics[width=\linewidth]{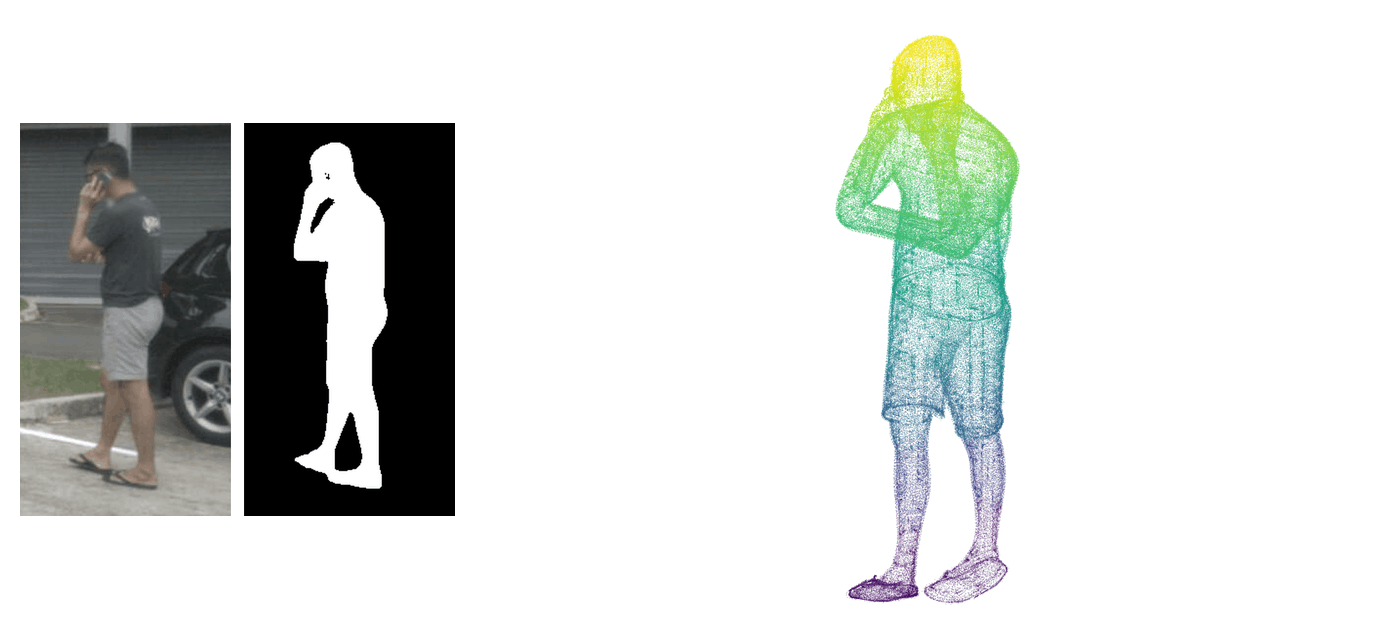}
    \end{minipage}

    \vspace{2pt}

    \begin{minipage}{0.49\textwidth}
        \centering
        \includegraphics[width=\linewidth]{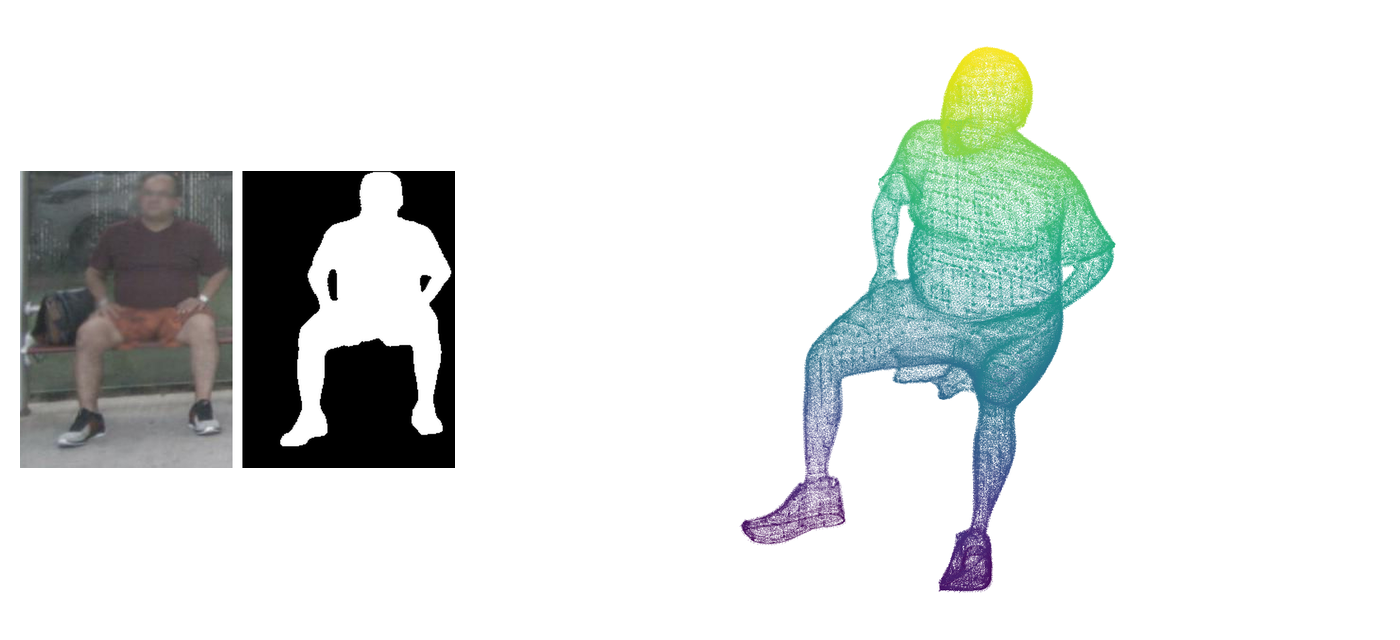}
    \end{minipage}
    \hfill
    \begin{minipage}{0.49\textwidth}
        \centering
        \includegraphics[width=\linewidth]{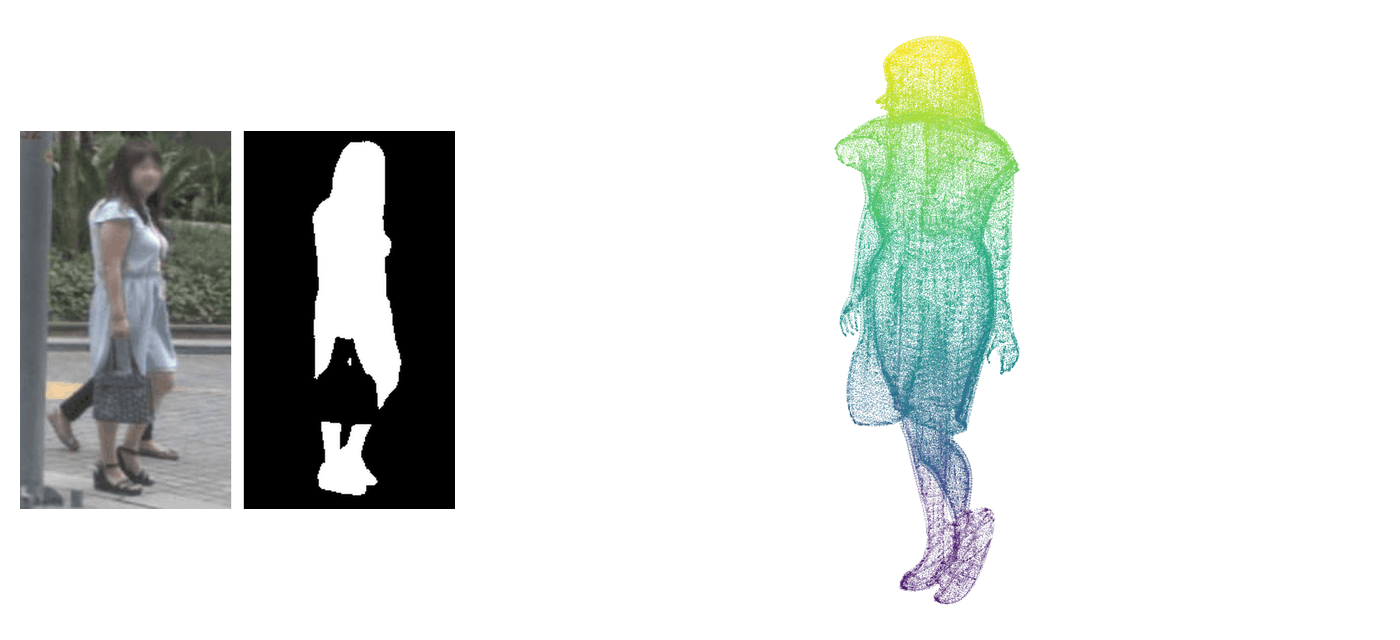}
    \end{minipage}

    \caption{Instance extraction and reconstruction by SAM3~\cite{carion2025sam3} and SAM3d~\cite{chen2025sam3d}.}
    \label{fig:instance}
\end{figure*}
nuCraftv2 builds upon the high-resolution occupancy formulation of nuCraft~\cite{zhu2024nucraft} and extends it along three orthogonal axes: the temporal frequency is raised from the $2\,\mathrm{Hz}$ key-frame rate of nuScenes annotations to the $20\,\mathrm{Hz}$ rate of raw LiDAR sweeps; every object that carries a 3D bounding-box annotation in nuScenes is modeled at the \emph{instance} level rather than 
only at the semantic level; and every voxel belonging to such an instance carries a \emph{panoptic} label $\ell = s\cdot 10^{3} + i$, where $s\in\{1,\ldots,10\}$ is one of the ten classes (barrier, bicycle, bus, car, construction\_vehicle, motorcycle, pedestrian, traffic\_cone, trailer, truck) and $i\in\mathbb{N}_{\geq 1}$ is a scene-unique instance id, while \emph{stuff} classes ($s\in\{11,\ldots,16\}$, e.g.\ drivable\_surface, sidewalk, terrain, manmade, vegetation) and empty voxels ($s{=}17$) all follow the convention $i{=}0$. The result is a high-\textbf{resolution} ($0.1/0.2/0.4\,\mathrm{m}$ voxels), high-\textbf{frequency} ($20\,\mathrm{Hz}$), high-\textbf{quality} (dense, instance-aware) occupancy dataset built on top of nuScenes, shown in Fig. \ref{fig:occ_compare_1} and Fig. \ref{fig:occ_compare_2}.
The data-curation pipeline is decomposed into two upstream pre-stages that compute detection and per-instance priors, and five core stages that fuse them into the final voxelized GT, executed in the following order, shown as Fig. \ref{Fig: nucraftv2_pipeline}.

\myparagraph{Cross-modal 3D detection.}
nuScenes provides GT 3D box annotations only at the $2\,\mathrm{Hz}$ key-frame rate, which is insufficient for the $20\,\mathrm{Hz}$ occupancy GT we target: between key frames, dynamic objects move by tens of centimeters and would otherwise leak into the static reconstruction.
We therefore run a cross-modal LiDAR–camera fusion detector~\cite{yan2023cmt} over every sweep, producing per-frame 3D detections at the full $20\,\mathrm{Hz}$ rate.
The cross-modal architecture is essential here, as image evidence helps recover small or partially-occluded objects that pure LiDAR detectors miss.
These dense detections complement the sparse key-frame GT and are used by later stages to filter out moving points before reconstruction.

\myparagraph{Per-instance 3D asset bank.}
A central design choice of nuCraftv2 is to model each foreground instance with a dedicated 3D asset, both to support the panoptic encoding $\ell = s\cdot 10^{3} + i$ and to compensate for the often sparse and occluded LiDAR returns on objects.
For every foreground annotation in a scene we sweep all six surround cameras across all samples and select the best observation by the lexicographic score $(\textit{is\_complete}, \textit{visibility\_level}, \textit{projection\_area})$, which favors a fully-visible, non-occluded, large-area view.
SAM3~\cite{carion2025sam3} then segments the cropped image with a class-conditioned text prompt drawn from a curated nuScenes-to-prompt table; only instances whose mask covers at least $20\%$ of the crop are retained, filtering out failures and ambiguous detections.
SAM3D~\cite{chen2025sam3d} lifts each surviving image–mask pair into a canonical-frame Gaussian-splatting reconstruction, producing a high-quality 3D template that is reused as a per-instance prior in the assembly stage. We present several examples of reconstructed instances, as illustrated in Fig.~\ref{fig:instance}.

\myparagraph{LiDAR-based pose re-estimation.}
The reason this re-estimation step is necessary is a subtle but consequential property of the nuScenes \texttt{ego\_pose} stream: the vertical component is \emph{identically zero} across the
entire dataset, treating every trajectory as perfectly planar.       
Reconstructions built directly on top of such poses collapse onto a flat manifold and inject a systematic vertical bias into the occupancy GT, particularly on sloped roads, overpasses, and multi-level parking lots where the true ego elevation can vary by several meters within a single sequence.
We therefore re-estimate per-frame poses with KISS-ICP, using the nuScenes poses only as the initial $xy$ seed and letting scan-to-map ICP recover the $z$-component along with the fine $\mathrm{SE}(3)$ refinement.         
The resulting estimates provide an accurate vertical reference that is propagated through all subsequent stages and is essential for producing geometrically faithful occupancy GT.

\myparagraph{Dynamic removal and panoptic label propagation.}
This stage produces a clean static background suitable for neural reconstruction while preserving panoptic semantics on every retained point.
For each sweep, we remove all points lying inside any GT-foreground box and any CMT detection box; the union is a strict super-set of the truly dynamic points and yields a conservative but high-precision background set that downstream reconstruction can trust.
Key-frame panoptic labels are inherited directly from the nuScenes panoptic annotations, while non-key frames within a five-frame window of the last key frame receive labels via $k{=}3$ KNN propagation in the global frame, with explicit scene-boundary checks to prevent label leakage across scenes. 
In parallel, we accumulate, for every annotated instance, its per-frame in-box points in the box-canonical frame, producing a multi-view canonical point cloud per instance that serves as a robust fallback whenever the SAM3D reconstruction quality is insufficient.

\myparagraph{Static-scene neural reconstruction.}
The cleaned background sweeps and the corrected KISS-ICP poses are fed into SHINE-Mapping~\cite{shine_mapping}, a sparse-hierarchical implicit field built on a feature octree decoded by shallow MLPs into both a signed distance value and a $32$-class semantic logit.
We train for $60$k iterations in batch mode jointly across the entire sequence, with a $0.3\,\mathrm{m}$ leaf voxel size, and extract a watertight mesh via marching cubes at $0.1\,\mathrm{m}$ resolution; the semantic head is queried at every vertex to label the mesh.
This yields both dense static geometry and consistent semantic labeling for the assembly stage.

\myparagraph{Per-frame voxelized panoptic occupancy assembly.}
The final stage materializes the $20\,\mathrm{Hz}$ occupancy GT by fusing the static reconstruction, the per-instance assets, and the BEV map prior at every LiDAR frame.
We first densely sample the post-processed semantic mesh into colored points and crop the result to the $[-51.2, 51.2]^{2}\!\times\![-5, 5]\,\mathrm{m}$ ROI in the current LiDAR frame after an alignment $\mathbf{T}_{\text{gt}} \mathbf{T}_{\text{kiss}}^{-1}$ that compensates for the residual KISS-ICP drift.
For each GT box in the frame, we fit the corresponding per-instance asset from the bank by anisotropically scaling its width, length, and height to exactly match those of the box, then aligning the scaled asset to the box pose via $\mathrm{SE}(3)$; sparse instances are additionally densified by NKSR~\cite{nksr} to mitigate majority-vote instability inside under-sampled voxels.
The union of points is then voxelized on the GPU via a one-hot scatter-add majority-vote scheme in a \emph{compact} panoptic-label space, avoiding the prohibitive memory footprint of a full     
one-hot tensor.
Finally, drivable-area voxels are overwritten according to the BEV map layout to inject \texttt{ped\_crossing}, \texttt{road\_divider}, and \texttt{lane\_divider} structures — using full masks for thin dividers and edge masks for area classes — producing the final sparse panoptic tensors at three resolutions.

\section{Additional Implementation Details}
\subsection{BEV Layout Acquisition Scheme}
\myparagraph{Universal Schema.} 
A BEV layout is a multi-hot image: no geometric alignment between
cameras, no LiDAR sweeps, no point-cloud completion. Any driving dataset
with HD-map annotations and 3D boxes can be projected into this shared
schema, which makes it straightforward to pool, filter and re-balance
training scenes across heterogeneous sources without re-labelling.

\myparagraph{User-editable.}
Because the layout is a raster image with interpretable channels, it can
be authored or edited directly: inserting a jaywalking pedestrian,
closing a lane, relocating a crosswalk, or swapping left-hand for
right-hand traffic all reduce to pixel edits. This is the mechanism by
which \MethodName supports \emph{arbitrary} layouts, including
cross-dataset and fully user-designed ones, as claimed in the abstract.

\myparagraph{Data engine.}
By decoupling scene {structure} (BEV layout) from scene
{appearance} (occupancy and downstream video), a single trained
OccDiT turns any stream of layouts into a stream of occupancy grids,
and, combined with our occupancy-based video module, a stream of
multi-view videos. This closes a practical data-engineering loop: rare
situations (long-tail intersections, construction zones, vulnerable-road-user
encounters) are expensive to {collect} but cheap to {author}
as BEV layouts, and \MethodName provides the generator that turns them into
usable training signal for downstream autonomous-driving models.

\subsection{Occupancy Generation Model}
\myparagraph{Architecture and Tokenization.}
The STOccDiT backbone consists of $L=24$ blocks with hidden dimension $D=768$, $12$ attention heads, and an FFN expansion factor of $4$. 
With stride-$8$ downsampling in the BEV encoder and a stride-$2$ patch embedding on the VAE latent, both BEV layout and occupancy tokens are mapped to a common $32 \times 32$ spatial grid, resulting in $S_b = S = 1024$ tokens per frame for each stream. 
Training is performed on sequences of $T=4$ frames. 
At inference time, we employ $30$ Euler discretization steps with classifier-free guidance at scale $w=2.0$.

\myparagraph{Training Procedure.}
To stabilize training and improve sample quality, STOccDiT is trained in two phases using the AdamW optimizer in bfloat16 automatic mixed precision. 
Both phases run for $500$ epochs with a base learning rate of $2 \times 10^{-4}$, weight decay of $10^{-3}$, gradient clipping of $1.0$, a $5$-epoch linear warmup, and cosine annealing down to $10^{-6}$.

In Phase 1, the temporal branch remains zero-gated and frozen. 
STOccDiT is trained as a BEV-conditioned single-frame model with classifier-free guidance dropout of $0.1$ and an exponential moving average (EMA) of decay $0.999$ activated from epoch $100$.

In Phase 2, the causal temporal layers—including temporal self-attention, its AdaLN parameters, and the per-frame embeddings—are activated and fine-tuned on $4$-frame sequences. 
A learning rate multiplier of $0.1$ is applied to all parameters inherited from Phase 1, while newly initialized temporal parameters use the full learning rate. 
Additional techniques include Gaussian noise augmentation with $\sigma=0.2$ on the clean stream and history dropout of $0.15$, while classifier-free guidance dropout remains at $0.1$. 
The EMA is disabled during this phase.

\myparagraph{Block Design.}
Each OccDiT block consists of three sub-layers modulated by AdaLN-Zero: spatial self-attention, causal temporal self-attention, and a feed-forward network.
In the spatial self-attention sub-layer, BEV layout tokens are concatenated with noisy occupancy tokens. 
A joint self-attention operation is performed over the combined sequence, allowing BEV features to be progressively refined. 
The clean stream uses the same attention weights but receives distinct AdaLN conditioning.

The causal temporal self-attention sub-layer interleaves clean and noisy tokens along the temporal axis. 
A causal mask ensures that a noisy token at time $t$ can attend only to clean tokens from previous frames and its own position. 
A learnable scalar gate initialized at zero controls the contribution of the temporal branch, allowing it to gradually activate during training.

A shared SwiGLU feed-forward network is applied independently to the noisy and clean streams under their respective AdaLN modulations. 
After the final block, only the noisy stream is passed through an unpatchify head to produce the velocity prediction. 
Spatial positions are encoded using 2D Rotary Position Embedding (RoPE), while temporal positions are represented by learnable per-frame embeddings. 
Timestep conditioning is injected via AdaLN, with the clean stream consistently conditioned on a zero timestep.

The dual-stream input combined with causal temporal masking enables efficient parallel training while ensuring strictly autoregressive generation at inference time.

\myparagraph{}{Flow Matching Objective.}
STOccDiT is trained using rectified flow matching in the latent space. 
For each frame $t$, a timestep $\tau_t$ is sampled from a logit-normal distribution. 
The noisy latent is constructed as
\begin{equation}
\mathbf{z}_t^{\tau_t} = (1 - \tau_t) \mathbf{z}_t + \tau_t \boldsymbol{\epsilon}_t,
\end{equation}
where $\boldsymbol{\epsilon}_t$ denotes standard Gaussian noise. 
The model predicts the velocity field $\mathbf{v}_\theta$ by minimizing a weighted mean squared error loss. 
A small-object-aware weighting scheme derived from the BEV layout is applied to improve reconstruction quality on rare classes such as pedestrians and bicycles.

\subsection{Driving Video Generation Model}

In this subsection, we present the training procedure and implementation details of the Geometry-Grounded View Expansion module and describe the design choices that enable GGVE to achieve cross-view consistency, high controllability, flexibility, and scalability.

\myparagraph{A unified mode-mixture training.}                                        
We train one Wan-based video diffusion backbone, augmented with a 
control branch that consumes the per-camera geometric signals. At each iteration the dataloader exposes a triplet of      
spatially adjacent cameras to the model, and we sample one of five                
training modes that partitions this triplet into a set of                         
\emph{anchor views}, which provide clean ground-truth pixels as                   
spatial anchors, and a set of \emph{target views}, which the model is             
asked to generate. Three of the five modes are pure text-to-video or image-to-video(which provide a first frame as reference for target views) and
use no anchor at all: they ask the model to generate one, two, or all
three of the adjacent cameras from text and geometry alone. The
remaining two modes are outpainting modes: one supplies a single
neighbouring camera as an anchor and asks the model to extend the
view to its adjacent partner, while the other supplies the two outer
cameras of the triplet as anchors and asks the model to outpaint the
middle camera between them. Together, the five modes cover every
useful combination of anchor count and target count under a triplet
of adjacent cameras, and ground every cross-view extension the model
will be asked to perform at inference time.
The latents of all active views---anchors and targets alike---are
concatenated, jointly noised at a single shared diffusion timestep,
and fed to the DiT as one token sequence; this keeps the main
backbone agnostic to the choice of mode and lets cross-view
self-attention operate uniformly across the whole rig. Mode
information enters the network only through the control branch,
which receives a per-view hint that is the clean ground-truth latent
at anchor positions and zero at target positions, together with a
binary spatial mask that flags anchor regions as observed. To prevent
the model from spending its capacity on the trivial task of copying
the anchor hint back into the denoiser output, the latent
reconstruction loss down-weights the anchor views by a small factor
and lets the target views dominate the gradient; target views are
what the model is forced to invent. A milestone schedule shifts the
mode mixture from text-to-video-heavy at the start of training
toward an approximately balanced distribution later on, so that the
model first acquires single-view fidelity and then progressively
shoulders more cross-view extrapolation. All five modes share the
same parameters, the same control branch, the same forward pass and
the same loss; the mode mixture is a curriculum on conditioning
patterns, not an ensemble of specialists.

\myparagraph{Composing the modes into surround synthesis.}
Given only a temporal occupancy volume---no reference image, no
reference video---we recover the full six-camera surround in four
sequential calls of the same model. The first call uses mode $(0,3)$
to jointly synthesise the front-left, front, and front-right views
from occupancy alone. The second and third calls each use mode
$(1,1)$, taking the previously generated front-left (resp.\
front-right) view as the conditioning anchor and outpainting the
back-left (resp.\ back-right) view at its correct relative pose. The
fourth call uses mode $(2,1)$, with the freshly generated back-left
and back-right views serving as conditioning anchors while the rear
view is outpainted between them. Cross-view consistency is not
enforced by sharing latents across calls; it emerges because every
call renders its semantic and coordinate buffers from the same
occupancy at the correct relative pose, so the generator solves a
geometrically over-constrained inverse problem at each step and the
only stable solution is one in which adjacent views agree. When a
single front-camera image is available at $t{=}0$, we attach it to
the mode-$(0,3)$ targets in the first call, turning that step into
image-to-video; the remaining chain is unchanged. The same procedure
therefore subsumes pure text-to-video surround generation and
start-frame-anchored generation under a single schedule.

\myparagraph{From sparse to dense camera rigs.}
The same outpainting modes generalise along the camera-count axis.
To insert a virtual camera between two existing ones, we place it at
the desired pose, render its semantic and coordinate buffers from the
same occupancy, and invoke mode $(1,1)$ or $(2,1)$ with the existing
views as conditioning anchors and the virtual view as the outpaint
target. Iterating this operation once doubles the rig; recursing
yields $6\!\to\!12\!\to\!24$ cameras over the same scene, with no
per-novel-view fine-tuning. Because every virtual camera is anchored
to occupancy through its rendered buffers, the densified rig remains
geometrically consistent in space and time, and the procedure
terminates at whatever camera density a downstream task requires.

\paragraph{Scalability.}
The end-to-end pipeline factorises as
\begin{equation}
\begin{array}{c}
\text{BEV} \\
\big\downarrow {\scriptstyle\text{any-BEV}\to\text{occ}} \\
\text{Occ}_{1:T} \\
\big\downarrow {\scriptstyle\text{render at }K\text{ poses}} \\
\{(\mathbf{s}_k,\mathbf{c}_k,\mathbf{p}_k)\}_{k=1}^{K} \\
\big\downarrow {\scriptstyle\text{view expansion}} \\
\mathbf{V}_{1:T,\,1:K}
\end{array}
\label{eq:ggve}
\end{equation}
where $\mathbf{s}_k$, $\mathbf{c}_k$ and $\mathbf{p}_k$ denote the
semantic buffer, coordinate buffer, and Pl\"ucker embedding of the
$k$-th camera. Reading Eq.~\eqref{eq:ggve} left-to-right, the video
model is reduced to the leaf of a chain whose intermediate stages
are explicit geometry: its input domain is occupancy and its task
is to map geometry to appearance. As a consequence, any BEV
layout---hand-drawn, retrieved, simulated, or perturbed---yields a
paired surround video without collecting new on-road footage; the
camera count $K$, the camera intrinsics and extrinsics, and the
temporal horizon $T$ are all set at query time without retraining;
and surround generation, image-anchored generation, and rig
densification share a single set of primitives, so new generation
patterns can be composed at inference without touching the model.
Geometry-grounded view expansion thus turns a video generator into an open-ended scene renderer driven entirely by BEV-level intent.

\myparagraph{GGEV module.}
The GGVE module is implemented based on Wan2.1-T2V-14B and initialized from its pretrained weights. For the ControlNet component, we employ a VACE-style ControlNet variant attached to the frozen Wan2.1-T2V-14B DiT backbone, which comprises 40 transformer blocks with a hidden dimension of 5120, 40 attention heads, and an FFN dimension of 13824. The control branch is a trainable side network with $L_c=8$ attention blocks, each structurally identical to the backbone DiT blocks, together with a $before\_proj$ linear layer at block 0 and a per-block $after\_proj$ output projection. The $after\_proj$ layers are zero-initialized so that, at the start of training, the controlled model exactly reproduces the pretrained backbone behavior.

Conditioning signals are encoded as follows. First, the target-view RGB latents, consisting of 16 channels from the Wan VAE, are concatenated channel-wise with the per-pixel Plücker camera embeddings. The Plücker embeddings have 6 channels and are downsampled by a 3D convolution. The concatenated features are then patchified by a $(1,2,2)$ Conv3d layer into dimension 5120 and zero-padded to match the backbone token length. Second, the semantic-buffer and coordinate-buffer latents are independently embedded by a dual-buffer patch embedder~\cite{infinicube} and added token-wise to the control stream. Within the control branch, block 0 fuses the main-branch hidden state via $c \leftarrow \mathrm{before\_proj}(c) + x$. The subsequent 8 control blocks share the same text and time conditioning $(\text{context}, t_\text{mod}, \text{freqs})$ as the main DiT and produce 8 hint tensors ${h_k}_{k=0}^{7}$.

The hint tensors are injected back into the main DiT at uniformly spaced layer indices $\mathcal{I}={0,5,10,15,20,25,30,35}$, corresponding to a stride of $\lfloor 40/8 \rfloor = 5$ and covering the full depth of the backbone. Specifically, hints are added through a scaled residual connection after each selected DiT block:
$$
x_\ell \leftarrow \text{DiTBlock}\ell(x{\ell-1}) + s \cdot h_{\pi(\ell)}, \qquad \ell \in \mathcal{I},
$$
where $s$ denotes the tunable injection scale, with default value $s=1$ during training, and $\pi$ maps each selected backbone layer index to its corresponding hint index.

During training, only the control-branch parameters are updated, while the DiT backbone, T5 text encoder, and VAE remain frozen. We train the model for 10k iterations at resolution $480\times832$ using 49 sampled frames. Optimization is performed with AdamW using a learning rate of $2\times10^{-5}$ and weight decay 0.01, together with bf16 mixed precision and gradient checkpointing. Training follows a curriculum over single-view, two-view, and three-view T2V settings, as well as outpainting modes.


\subsection{Evaluation Metrics Details}
Inspired by X-Scene~\cite{x-scene} and UniScene~\cite{uniscene}, we adopt the following evaluation metrics to assess the quality of occupancy generation shown in Table \ref{tab:occ_gen}.
\textbf{(i) 3D Volume} accumulates a $21{\times}21$ voxel-level confusion
matrix over the entire validation split; {mIoU} averages IoU across
the 20 non-free semantic classes and {IoU} is the binary
occupied-vs-free (geometric) IoU.
\textbf{(ii) BEV Top-down} performs the same accounting after projecting
both volumes to a 2D label map (lowest-$z$ non-free class per column).
\textbf{(iii) BEV vs.\ Layout} compares the per-channel BEV {layout}
(the conditioning input) against the top-down projection of the
{predicted} occupancy across the 15 layout channels.
\textbf{(iv) Cam Render} feeds per-camera depth/semantic PNGs of GT and
prediction through InceptionV3~\cite{InceptionV3} to obtain FID(Fréchet Inception Distance) and KID (Kernel Inception Distance) on the depth and semantic streams, reported as their mean.
\textbf{(v) BEV Render} colour-maps the BEV semantic and BEV ``height''
projections on-the-fly and feeds them through the same FID and KID pipeline.
These five evaluation families are all derive from a
single per-frame $(256{\times}256{\times}25)$ predicted occupancy and its
ground truth.

\section{Additional Qualitative Results}
To complement the qualitative results in the main paper, we provide six
groups of additional visualizations that together illustrate every key
capability of our framework: dense temporal control, cross-view consistency,
trajectory editing, image-anchored surround generation, geometry-grounded
view expansion, resulting reconstruction quality and the scene editing. 

\begin{figure*}[t]
    \centering
    \includegraphics[width= 0.9\textwidth]{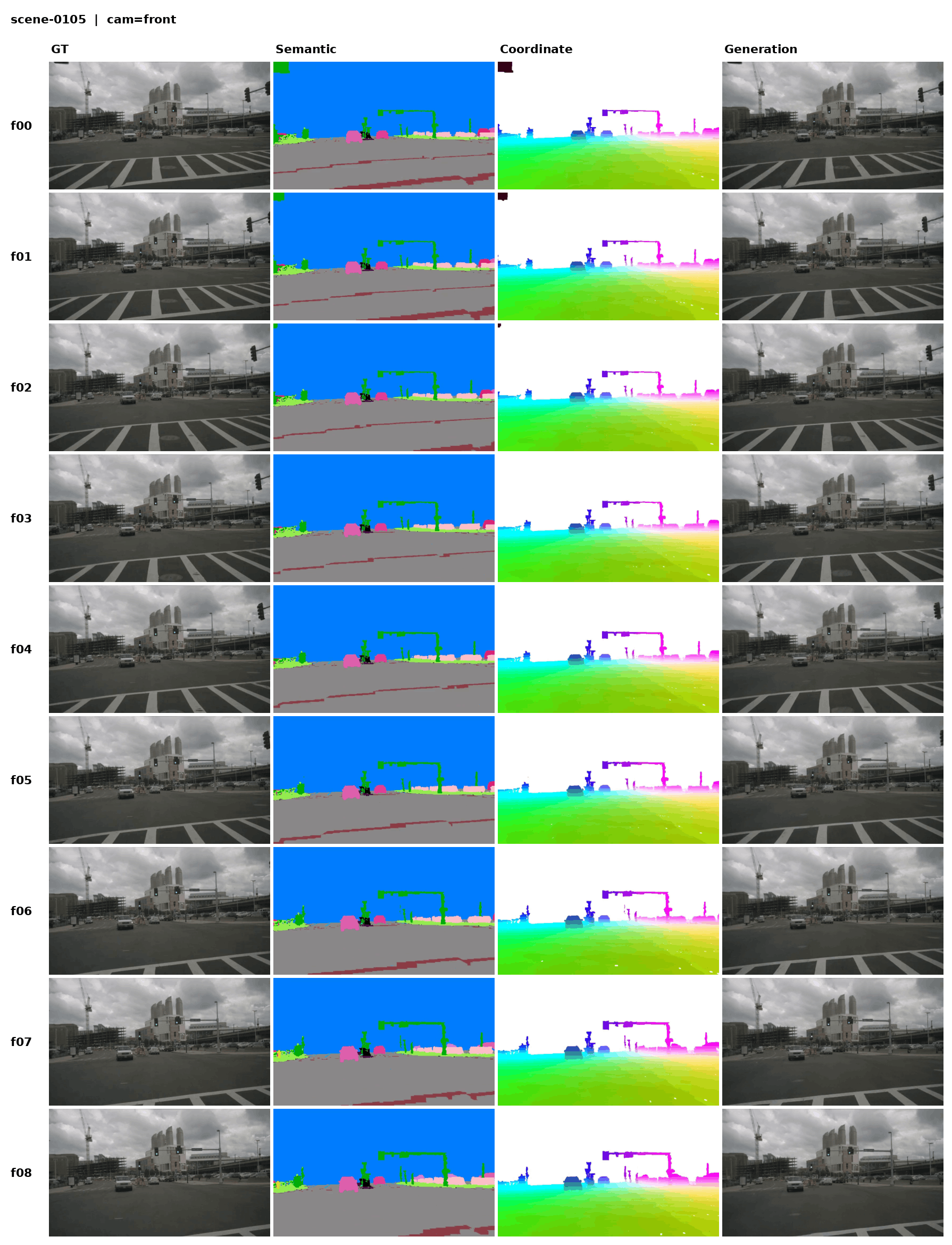}
    \caption{\textbf{Per-frame condition control (I2V).}
Each row is one timestep; columns from left to right:
GT, $12\,\mathrm{Hz}$ semantic buffer, $12\,\mathrm{Hz}$ coordinate buffer, our generation.
Generation tracks the dense control signal pixel-for-pixel.}
    \label{fig:add_vis_1}
\end{figure*}

\myparagraph{Per-frame condition control under the I2V setting.}
Fig.~\ref{fig:add_vis_1} probes how faithfully the generated video
follows the per-frame condition stream. For each case we visualize the
$12$\,Hz semantic and coordinate buffers next to the corresponding generated
front-camera frame. Two observations stand out: (i) the high-frequency
control signal is dense and temporally precise, with no blur or
inter-frame drift in the conditioning channel, and (ii) the generation
tracks every condition transition tightly --- structural cues such as lane
boundaries, dynamic agents, and the ego-anchored coordinate frame are
reproduced pixel-for-pixel, which is the prerequisite for any downstream
controllable use of the model.

\begin{figure*}[t]
    \centering
    \includegraphics[width=\textwidth]{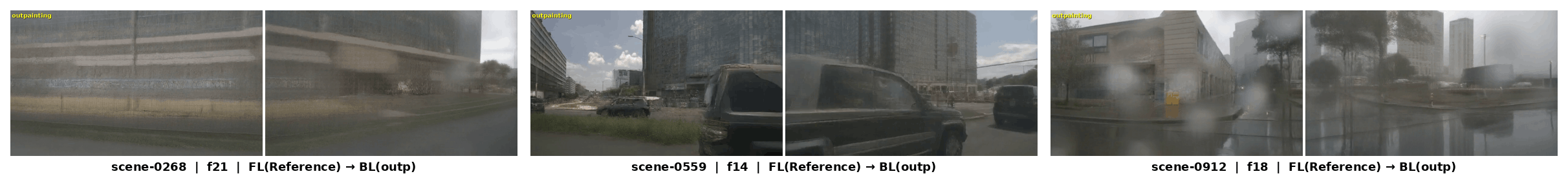}\\[2pt]
    \includegraphics[width=\textwidth]{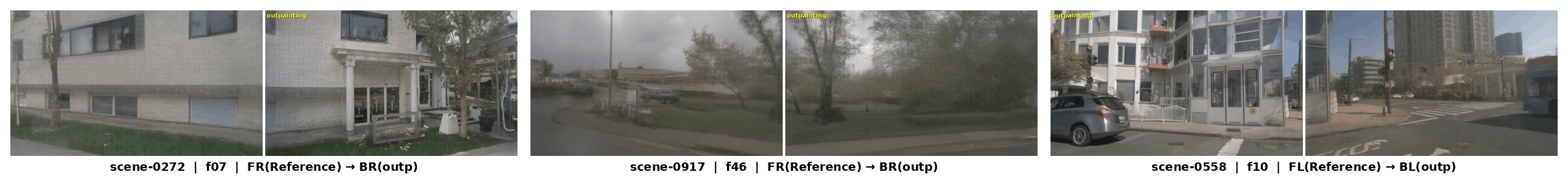}\\[2pt]
    \includegraphics[width=\textwidth]{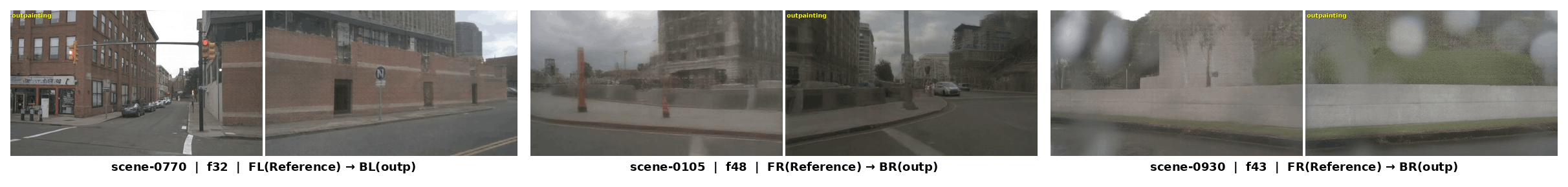}\\[2pt]
    \includegraphics[width=\textwidth]{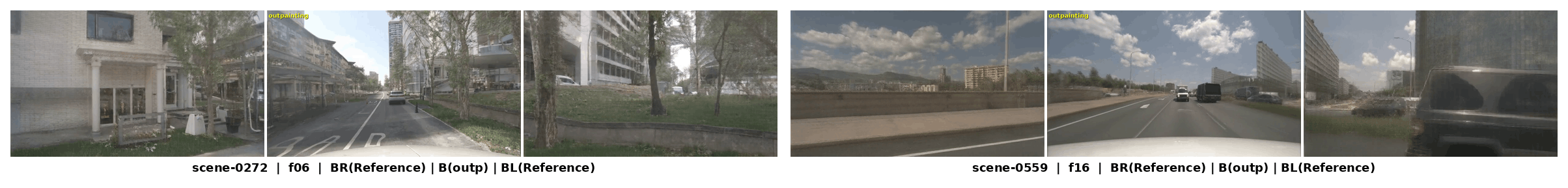}\\[2pt]
    \includegraphics[width=\textwidth]{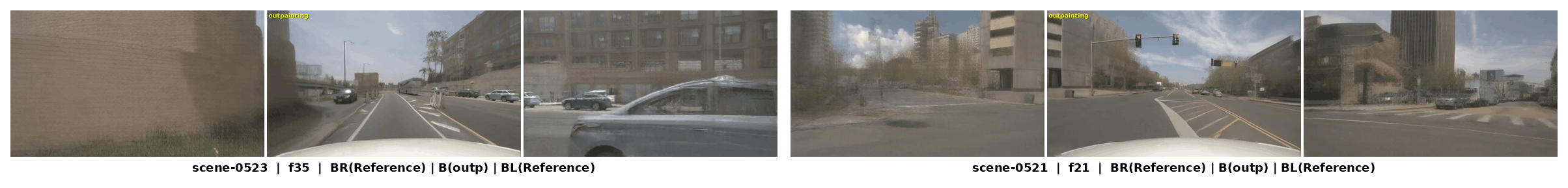}
    \caption{\textbf{Cross-view consistency of $2$- / $3$-view outpainting.}
Each row holds three $2$-view (top) or two $3$-view (bottom) cases.
The \texttt{outp} view is generated from its condition buffer alone and
fuses seamlessly with the reference view(s).}
    \label{fig:add_vis_2}
\end{figure*}

\begin{figure*}[t]
    \centering
    \includegraphics[width=\textwidth]{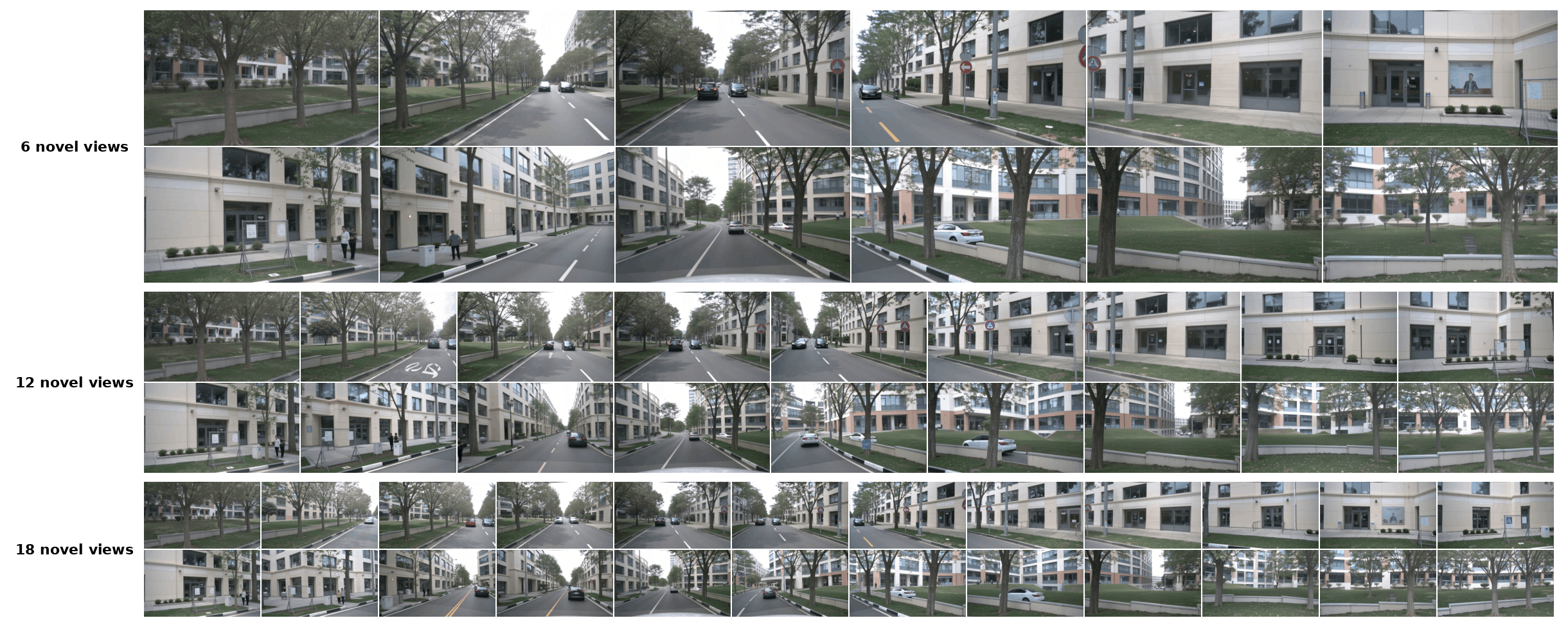}\\[2pt]
    \includegraphics[width=\textwidth]{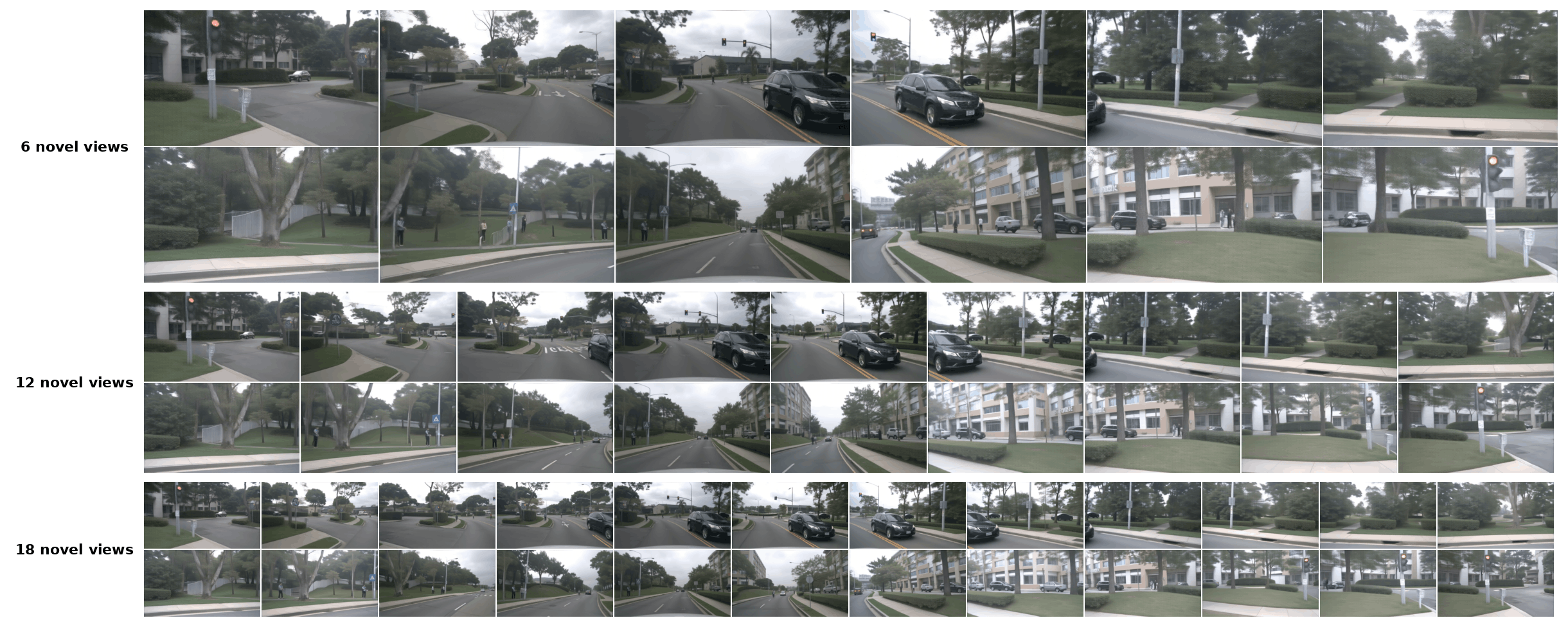}\\[2pt]
    \includegraphics[width=\textwidth]{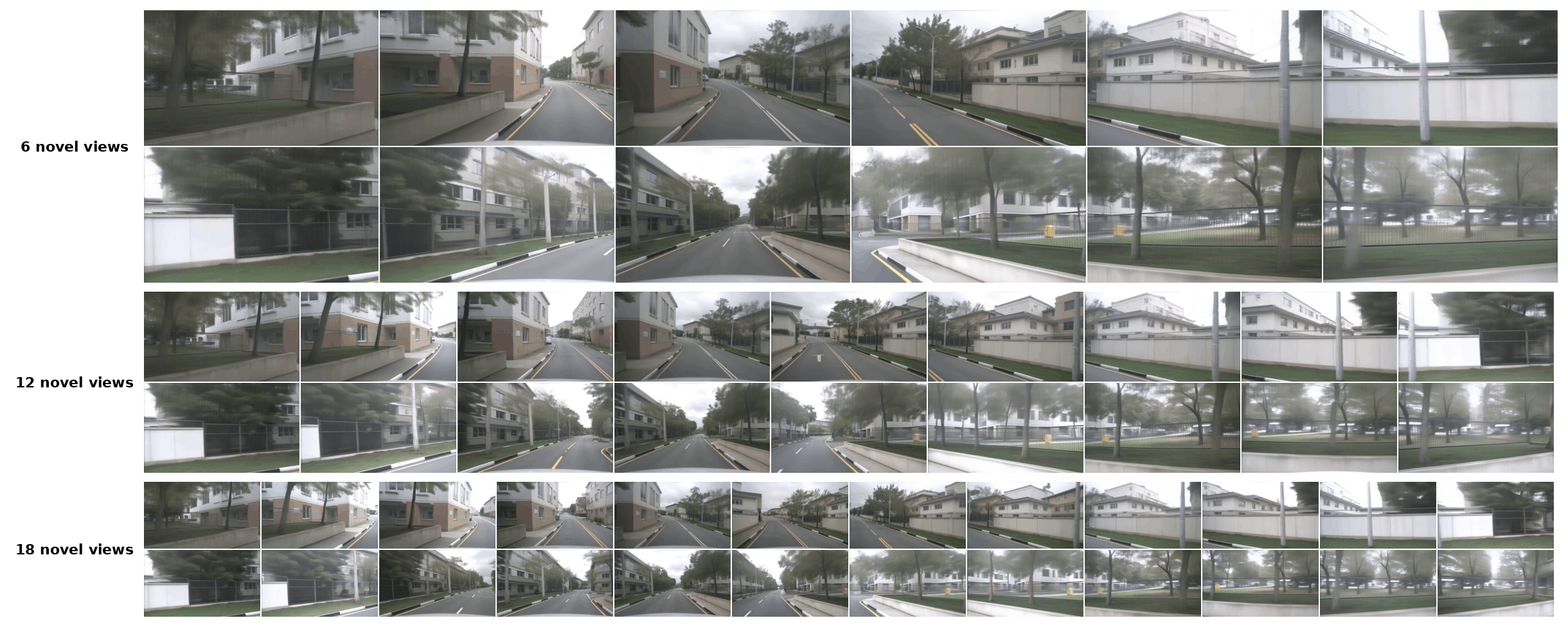}
    \caption{\textbf{Geometry-grounded view expansion.}
Per row, top to bottom: $12$-, $18$-, $24$-view rigs ($6$ baseline +
$6$/$12$/$18$ novel cameras). The procedure imposes no upper bound on
the novel-camera count.}
    \label{fig:add_vis_5}
\end{figure*}

\myparagraph{Cross-view consistency via $2$- and $3$-view outpainting.}
Fig.~\ref{fig:add_vis_2} illustrates the outpainting capability across
both the $2$-view and $3$-view configurations. Given the reference view(s)
and the per-frame condition buffer for the held-out view, our model
synthesizes the missing view in a way that is internally consistent --- the
shared road surface, building facades and overlapping objects align across
adjacent cameras within a row, with no visible seam, content drift, or
appearance mismatch. This cross-view consistency is what later enables the
chained, arbitrary-length view expansion shown in Fig.~\ref{fig:add_vis_5}.

\begin{figure*}[t]
    \centering
    \includegraphics[width=\textwidth]{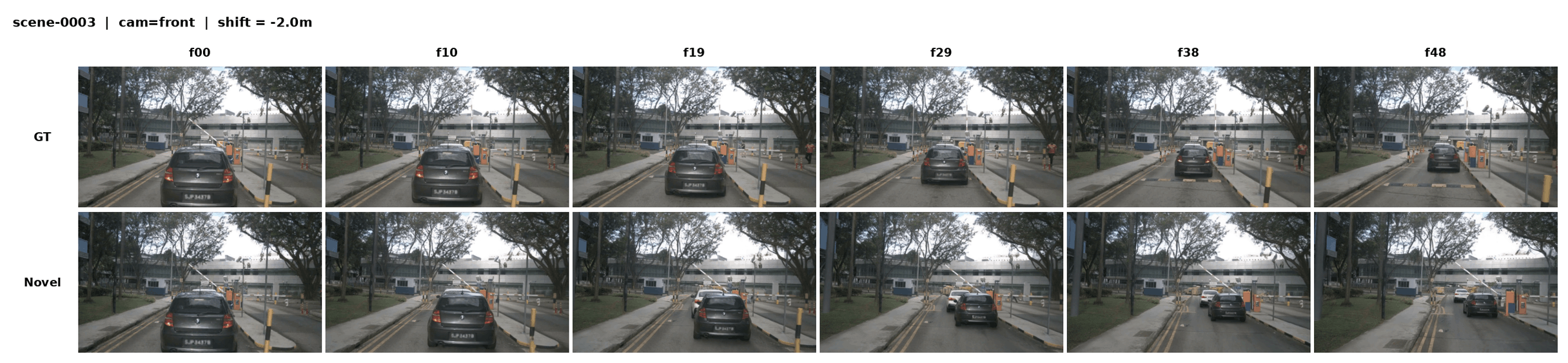}\\[2pt]
    \includegraphics[width=\textwidth]{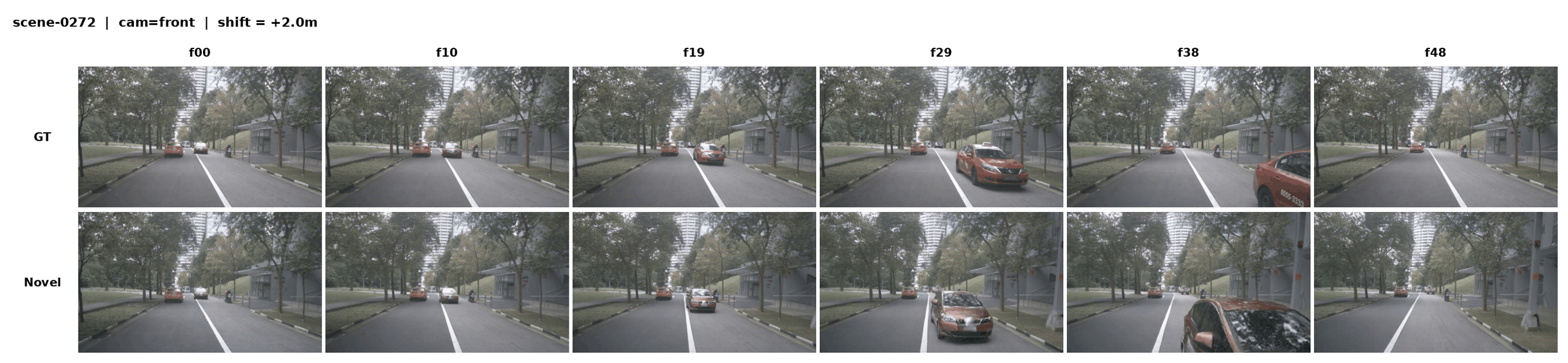}\\[2pt]
    \includegraphics[width=\textwidth]{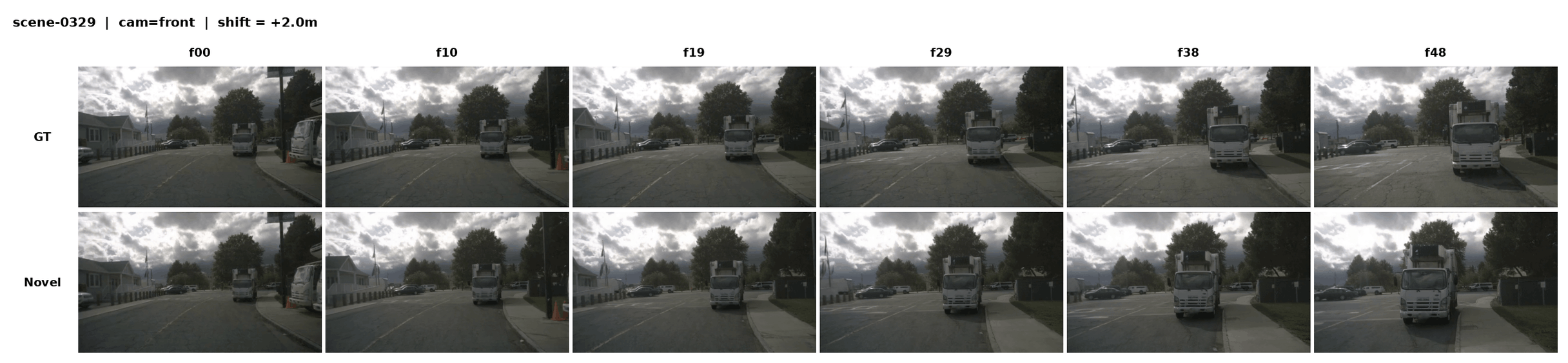}\\[2pt]
    \includegraphics[width=\textwidth]{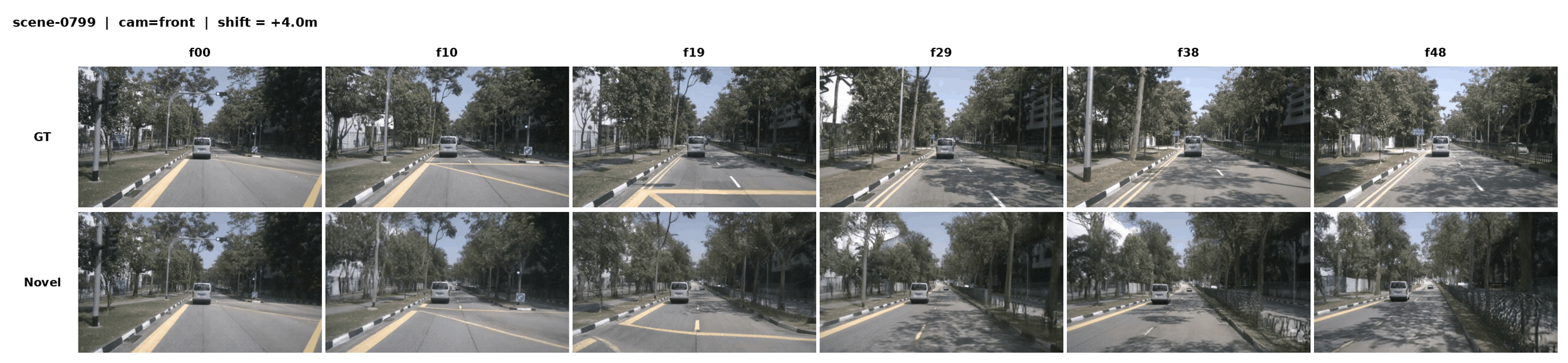}\\[2pt]
    \includegraphics[width=\textwidth]{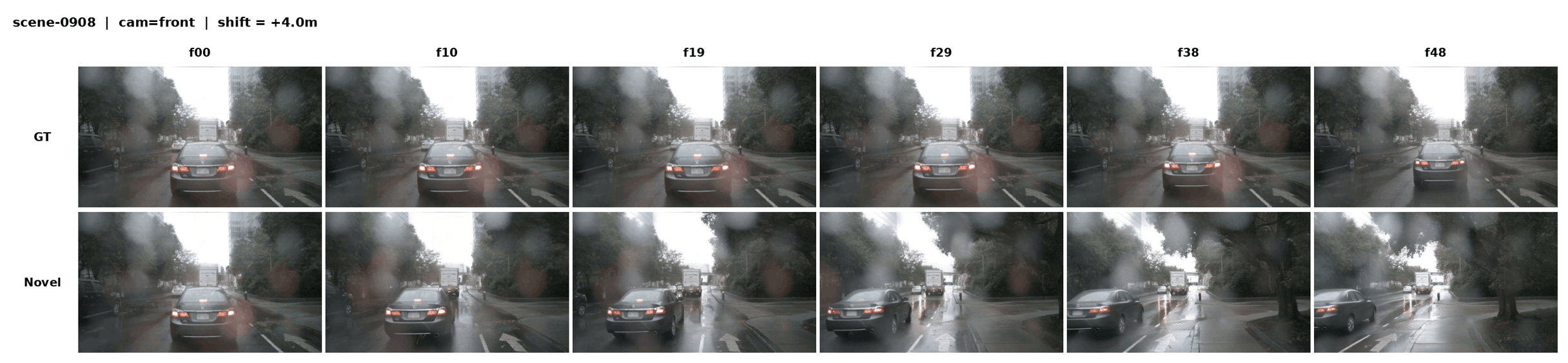}
    \caption{\textbf{Novel-trajectory generation.}
Top: GT trajectory. Bottom: novel trajectory after laterally shifting the
ego path by the amount in the title ($\pm 2$\,/\,$\pm 4$\,m).
Generation stays photorealistic across the full shift range.}
    \label{fig:add_vis_3}
\end{figure*}

\myparagraph{Novel-trajectory generation via occupancy resampling.}
Fig.~\ref{fig:add_vis_3} shows that, under the I2V setting, our model can
generate front-camera videos along arbitrary novel trajectories simply by
resampling the occupancy grid along a shifted ego path before re-rendering
the condition buffers. We compare the original GT trajectory against four
novel offsets ($\pm 2$\,m and $\pm 4$\,m lateral shift) and observe that
each generated video remains photorealistic, temporally smooth, and
geometrically consistent with the perturbed ego path --- evidence that the
trajectory-editing pipeline is decoupled from any specific recorded
trajectory and can be steered freely at inference time.

\begin{figure*}[t]
    \centering
    \includegraphics[width=\textwidth]{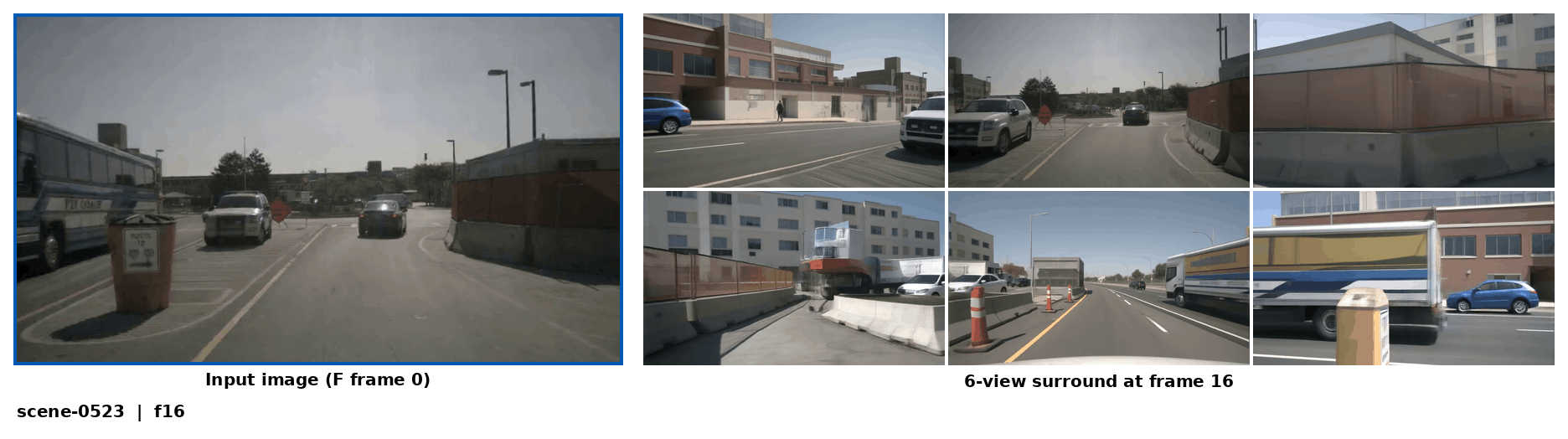}\\[2pt]
    \includegraphics[width=\textwidth]{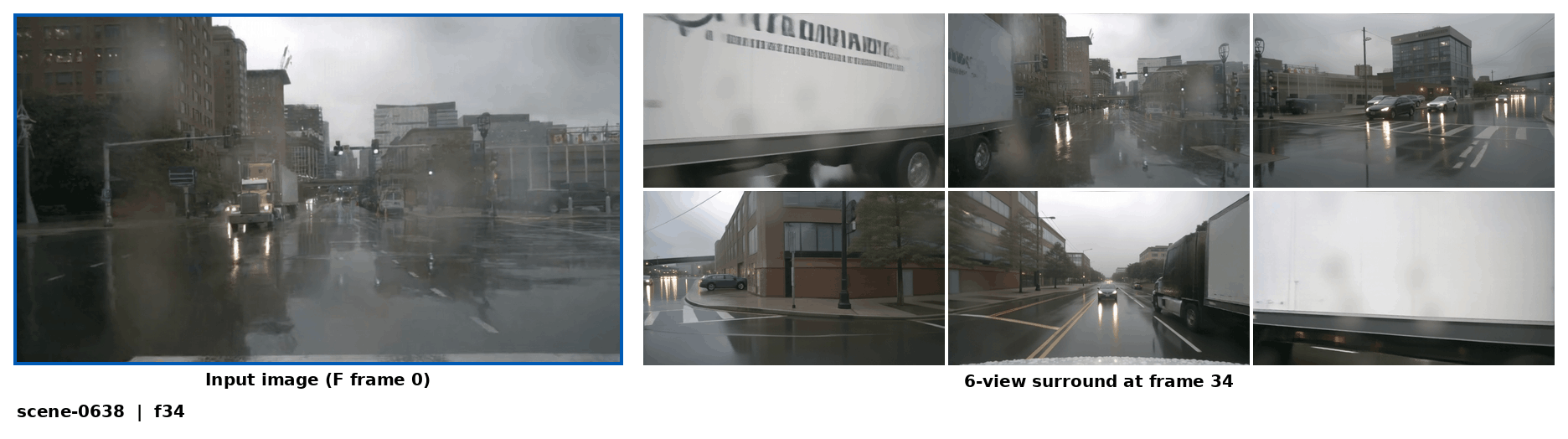}\\[2pt]
    \includegraphics[width=\textwidth]{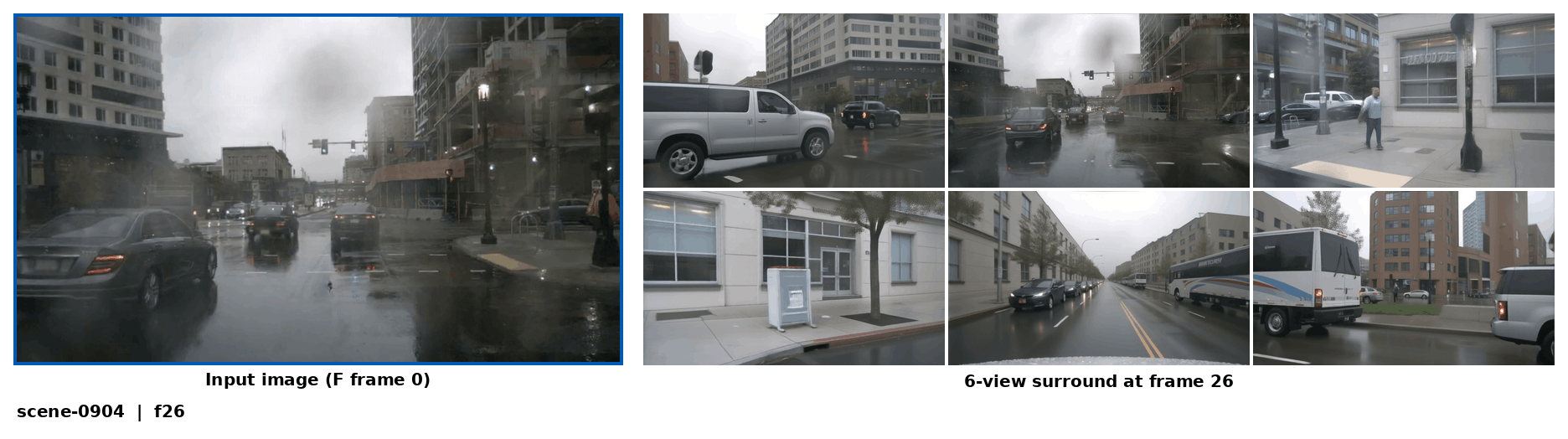}\\[2pt]
    \includegraphics[width=\textwidth]{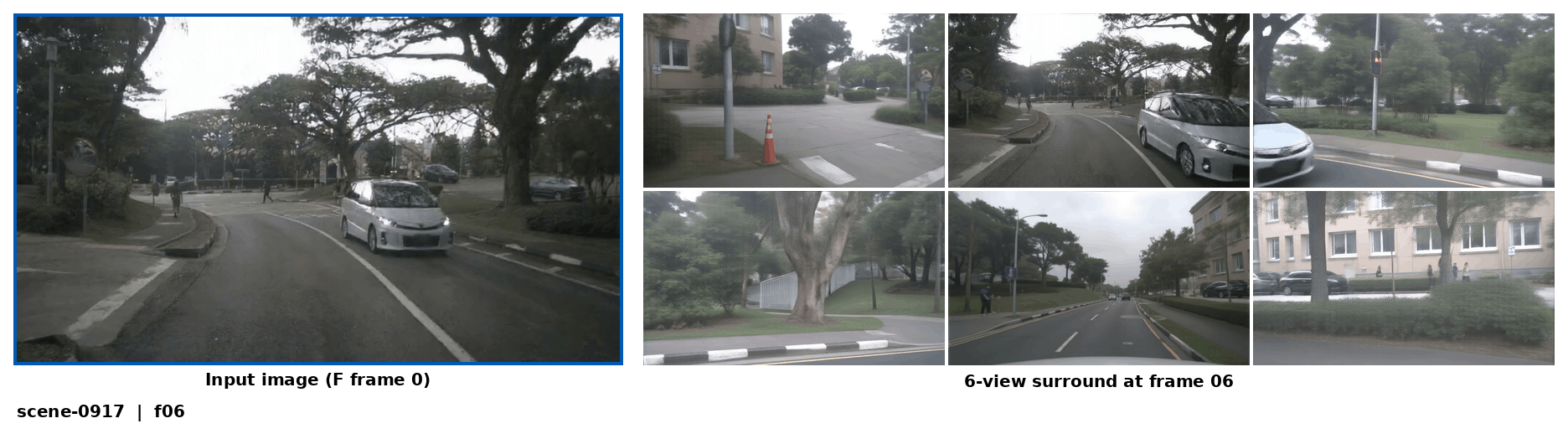}
    \caption{\textbf{Anchor-image surround generation.}
Given the $t{=}0$ front frame as the only anchor (left), our model
synthesizes the full $2{\times}3$ surround rig}
    \label{fig:add_vis_4}
\end{figure*}

\myparagraph{Anchor-image surround generation.}
Fig.~\ref{fig:add_vis_4} demonstrates that, beyond generating surround-view
videos directly from an occupancy grid, we can also condition the entire
$6$-view rig on a single anchor image (e.g., the front-camera frame at
$t{=}0$). For each case we show the anchor on the left and the
$2{\times}3$ surround output at a randomly sampled frame on the right.
The generated front, front-left, front-right, back, back-left, and
back-right cameras share illumination, scene layout, and dynamic agents
consistent with the anchor, providing a flexible "single-image-to-surround"
workflow that complements the standard occupancy-driven pipeline.

\myparagraph{Novel view expansion: from $6$ to infinite novel rigs.}
Fig.~\ref{fig:add_vis_5} demonstrates that geometry-grounded view
expansion scales to an arbitrary number of novel cameras without retraining.
Starting from the $6$ baseline cameras (FL, F, FR, BR, B, BL), we
progressively add $6$, $12$, and $18$ novel cameras interpolated between
adjacent baseline pairs, yielding $12$-, $18$-, and $24$-view rigs.
Each row preserves the $6$ original baseline cameras (highlighted) and
adds outpainted novel views that are spatially consistent with their
neighbours. The expansion procedure is purely a sampling decision in the
camera-rig space: there is no upper bound, and the same machinery would
densify the rig further to dozens of novel cameras when needed for downstream
sparse-to-dense applications.

\begin{figure*}[t]
    \centering
    \includegraphics[width=\textwidth]{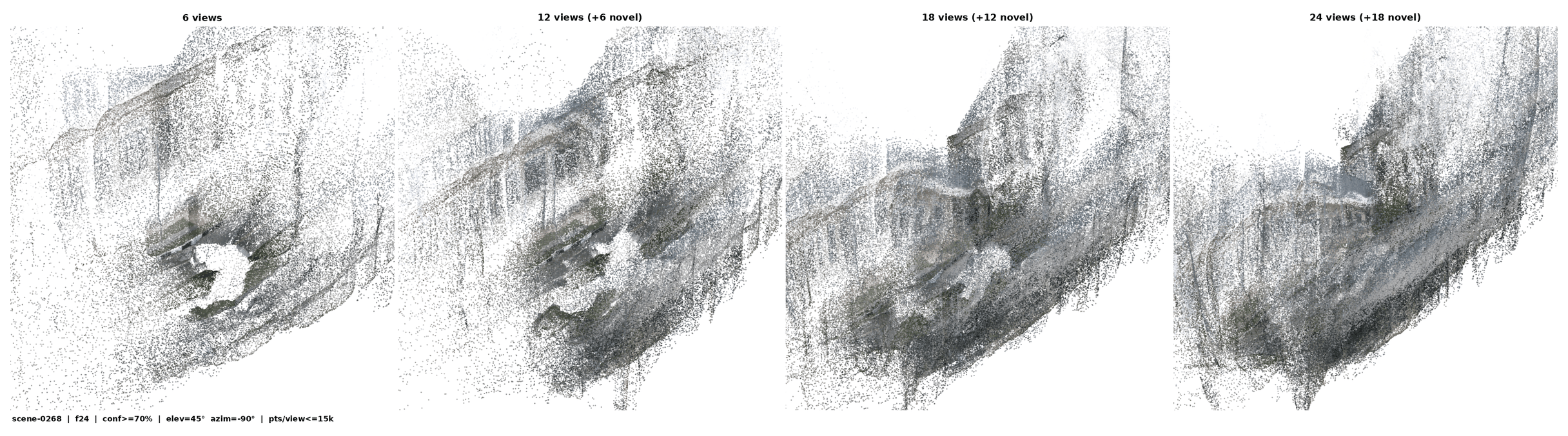}\\[2pt]
    \includegraphics[width=\textwidth]{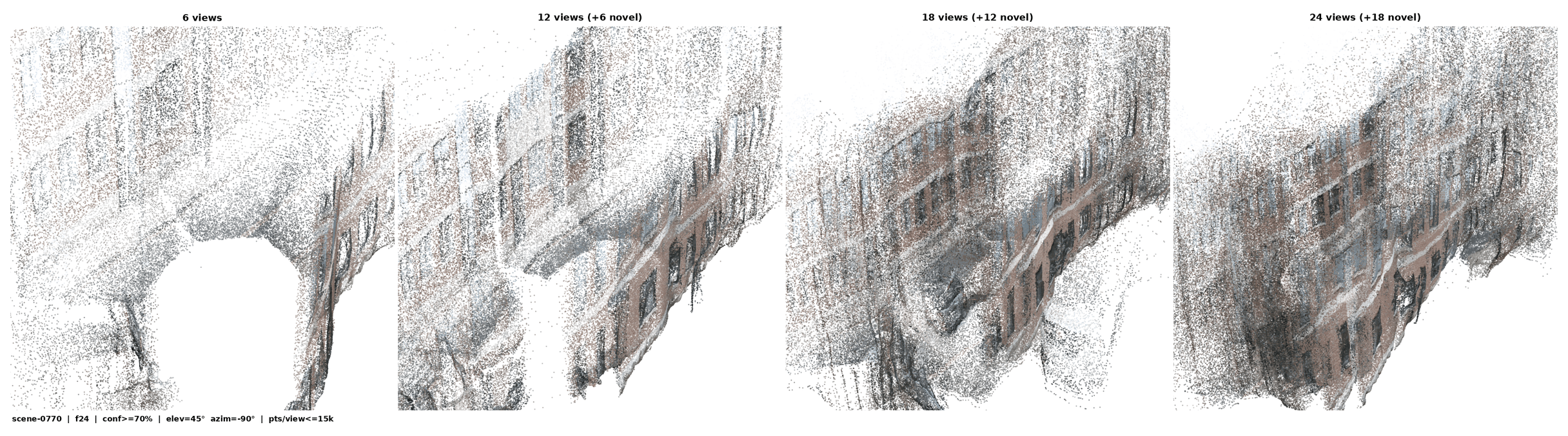}\\[2pt]
    \includegraphics[width=\textwidth]{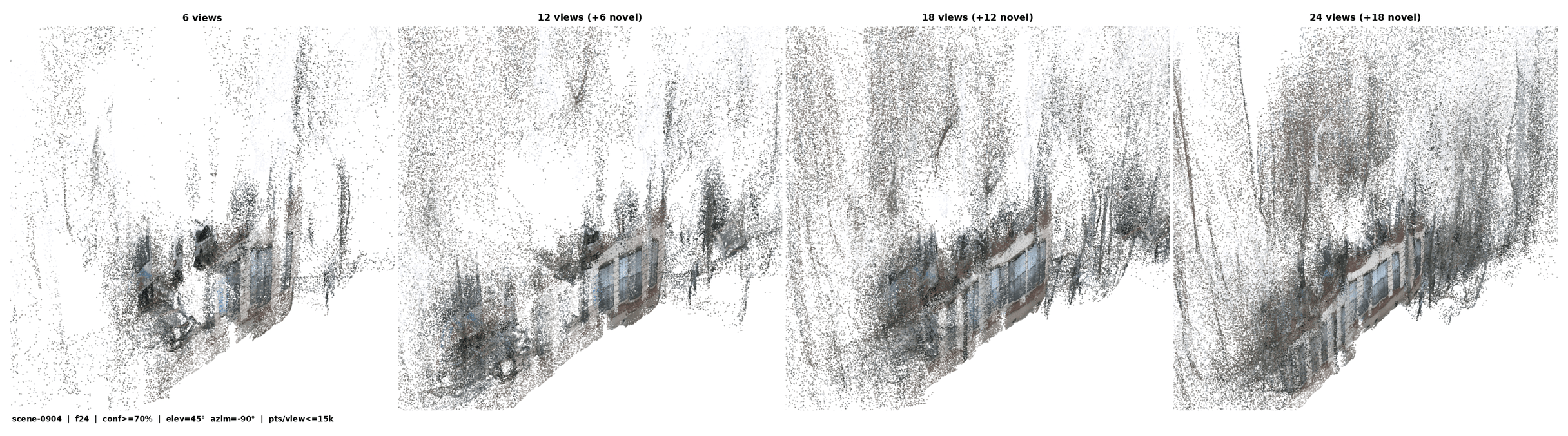}\\[2pt]
    \includegraphics[width=\textwidth]{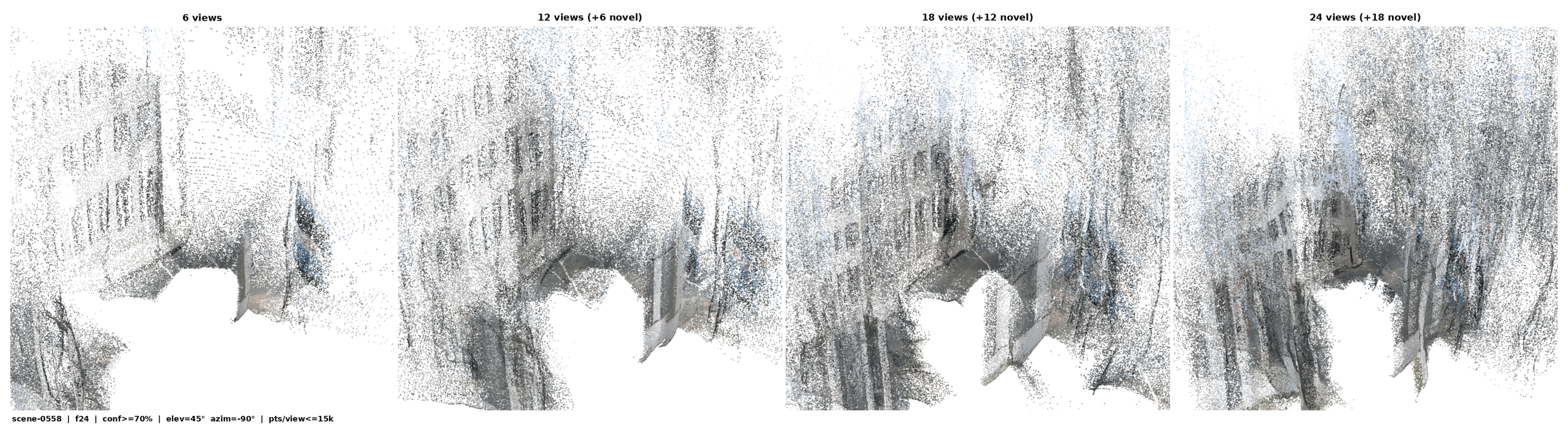}
    \caption{\textbf{More views $\Rightarrow$ denser reconstruction.}
$6$/$12$/$18$/$24$-view sets fed into VGGT~\cite{wang2025vggt} and rendered from
one fixed chase-camera viewpoint. Density grows monotonically with view
count.}
    \label{fig:add_vis_6}
\end{figure*}

\myparagraph{More views $\Rightarrow$ denser reconstruction.}
Fig.~\ref{fig:add_vis_6} closes the loop by quantifying, in $3$D, the
benefit of the additional novel views from~Fig.~\ref{fig:add_vis_5}.
For each scene we feed the $6/12/18/24$-view set into VGGT~\cite{wang2025vggt} and render the
reconstructed point cloud from the same fixed chase-camera viewpoint with
identical confidence, depth, and spatial filters across all four panels.
The resulting clouds become visibly denser as more novel views are added:
the road surface fills in, distant buildings gain structure, and
previously sparse regions (e.g., side curbs, shop fronts) acquire complete
coverage. This confirms that our outpainted novel cameras are not
geometrically vacuous --- they contribute genuine $3$D evidence that
state-of-the-art geometry foundation models can immediately exploit.

\begin{figure*}[t]
    \centering
    \includegraphics[width=\textwidth]{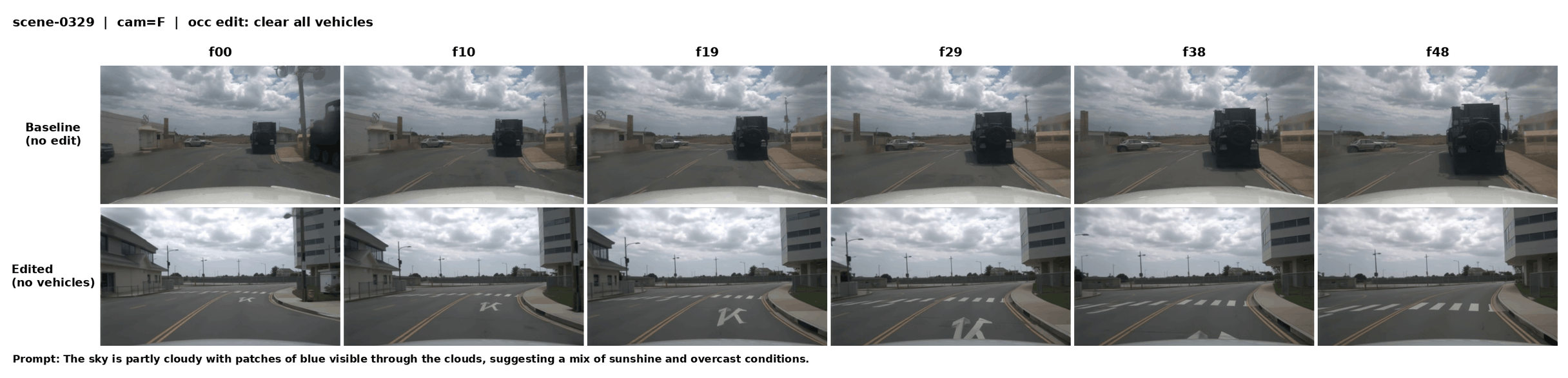}\\[2pt]
    \includegraphics[width=\textwidth]{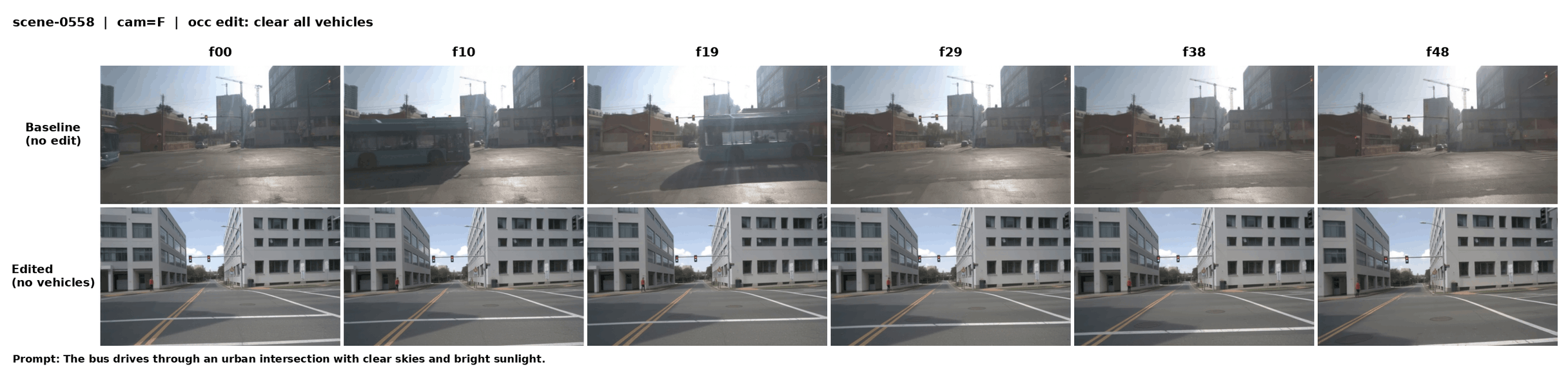}\\[2pt]
    \includegraphics[width=\textwidth]{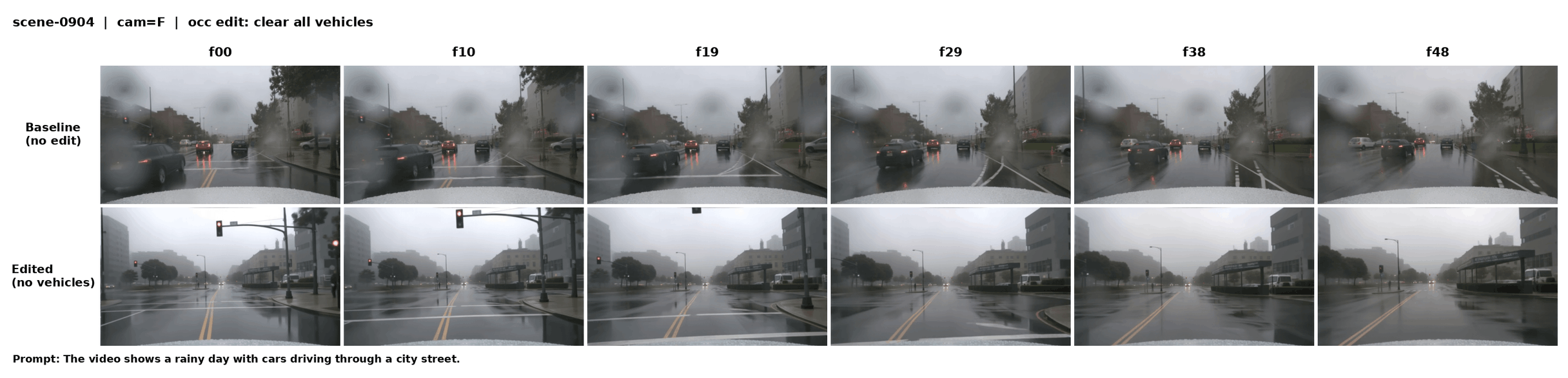}\\[2pt]
    \includegraphics[width=\textwidth]{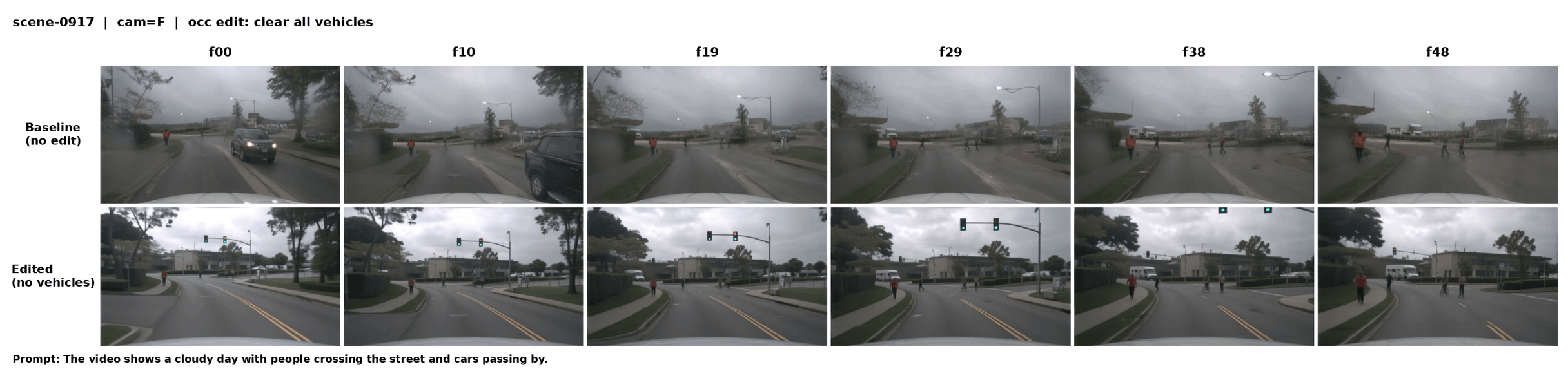}
    \caption{\textbf{Occupancy editing — vehicle clearing.}
Top: unedited baseline. Bottom: generation after clearing all vehicle
voxels from the input occupancy.}
    \label{fig:add_vis_7}
\end{figure*}

\myparagraph{Occupancy editing.}
The GGVE model takes semantic and coordinate buffer rendered from occupancy as control signal, so edits at the occupancy level propagate directly into the rendered buffers. We demonstrate this by removing all dynamic vehicles from the occupancy grid and re-generate the scene. As shown in Fig.~\ref{fig:add_vis_7}, the cleared vehicles disappear cleanly while the structure of surrounding static scene and ego trajectory remain unchanged, evidencing that occupancy-level edits translate to controllable, localised modifications in pixel space.

\begin{figure*}[t]
    \centering
    \includegraphics[width=0.9\textwidth]{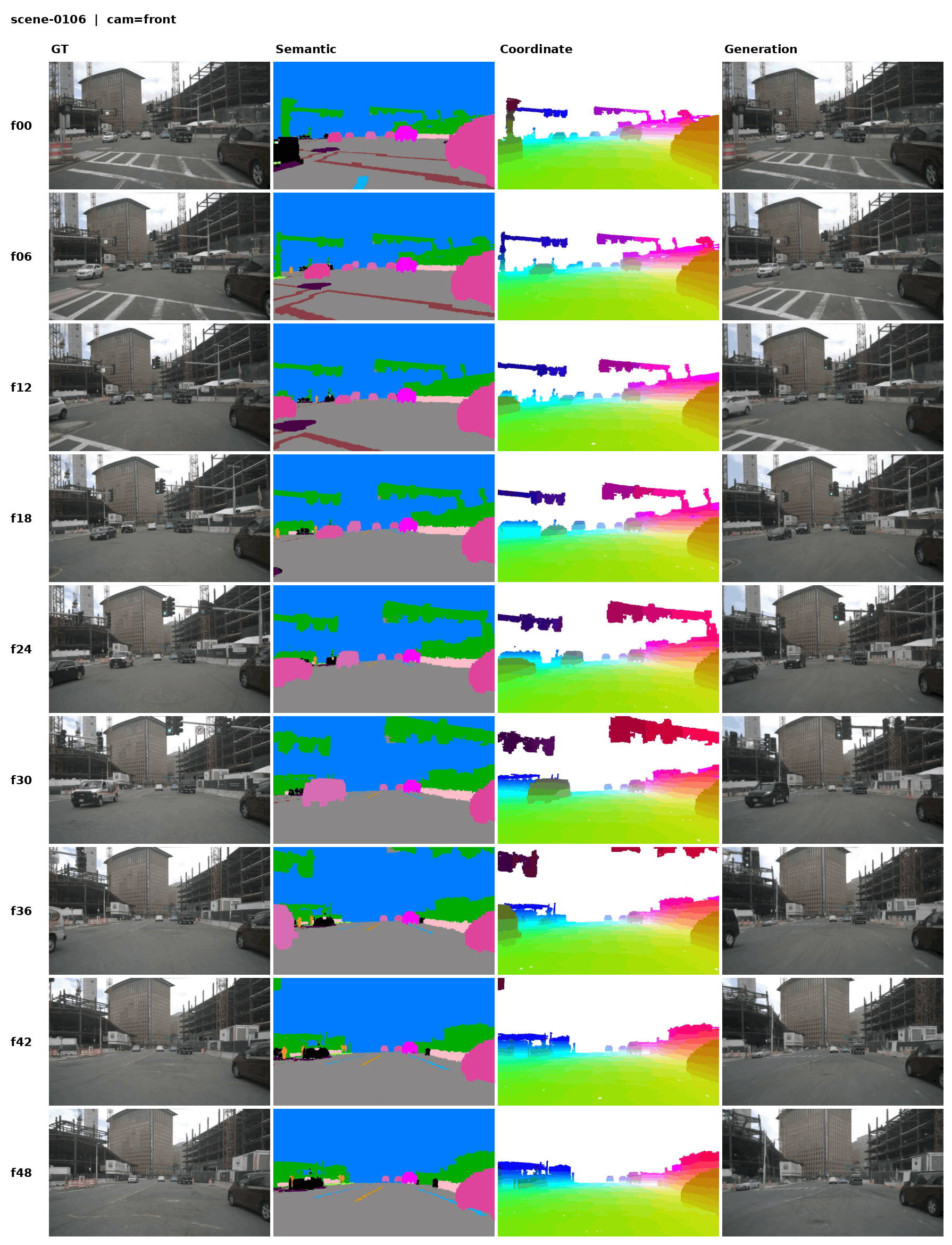}
    \caption{\textbf{Per-frame condition control (I2V).}The generated video maintains precise controllability across long frame sequences.}
    \label{fig:add_vis_1_1}
\end{figure*}

\end{document}